\documentclass{article}

\PassOptionsToPackage{numbers, compress}{natbib}
\usepackage{amsmath}
\usepackage{microtype}
\usepackage{graphicx}
\usepackage{subfigure}
\usepackage{hyperref}
\hypersetup{
    colorlinks=true,
    linkcolor=blue,
    filecolor=magenta,      
    urlcolor=cyan,
    citecolor=black
}
\usepackage{booktabs} 
\usepackage{amsfonts}
\usepackage{bm}
\usepackage{threeparttable}
\usepackage{tablefootnote}

\usepackage{lipsum}
\usepackage{comment}

\usepackage{multirow}
\usepackage[usenames, dvipsnames]{color}
\usepackage{amsmath}

\usepackage{hyperref}

\usepackage{enumitem}

\usepackage{mathtools}

\usepackage{listings}

\usepackage[english]{babel}
\usepackage[utf8x]{inputenc}

\usepackage{listings}
\usepackage{color}

\definecolor{mygreen}{rgb}{0,0.6,0}
\definecolor{mygray}{rgb}{0.5,0.5,0.5}
\definecolor{mymauve}{rgb}{0.58,0,0.82}

\usepackage{marginnote}
\usepackage{soul} 

\usepackage[colorinlistoftodos]{todonotes}

\newcommand{\ignore}[1]{}

\newcommand{\code}{\texttt}                             



     \usepackage[final]{neurips_2020}

\usepackage[T1]{fontenc}    
\usepackage{hyperref}       
\usepackage{url}            
\usepackage{booktabs}       
\usepackage{amsfonts}       
\usepackage{nicefrac}       
\usepackage{microtype}      
\usepackage{amsmath}
\usepackage{amssymb}

\usepackage[usenames, dvipsnames]{color}

\usepackage{tabularx}
\usepackage{multirow}
\usepackage{hhline}

\usepackage{natbib}
\usepackage{algcompatible}
\usepackage{algorithm}
\usepackage{enumitem}
\usepackage{multicol}

\usepackage{titlesec}
\titlespacing\section{0pt}{0pt plus 0pt minus 1pt}{0pt plus 0pt minus 1pt}
\titlespacing\subsection{0pt}{0pt plus 0pt minus 1pt}{0pt plus 0pt minus 1pt}
\titlespacing\subsubsection{0pt}{0pt plus 0pt minus 1pt}{0pt plus 0pt minus 1pt}

\newcommand{\lstm}{\textsc{lstm}}

\newtheorem{remark}{Remark}

\def\vb{{\bm{b}}}
\def\vc{{\bm{c}}}

\def\vh{{\bm{h}}}
\def\vi{{\bm{i}}}

\def\vm{{\bm{m}}}

\def\vo{{\bm{o}}}
\def\vp{{\bm{p}}}

\def\vr{{\bm{r}}}

\def\vu{{\bm{u}}}

\def\vx{{\bm{x}}}
\def\vy{{\bm{y}}}

\def \RR {{\mathbb{R}}}

\def \bx {{\bm x}}

\def \bv {{\bm v}}
\def \bv {\mathbf{v}}

\def \Ab {{\mathbf{A}}}

\def \Db {{\mathbf{D}}}
\def \Ub {{\mathbf{U}}}
\def \Wb {{\mathbf{W}}}

\usepackage{verbatim}

\usepackage{wrapfig}

\title{
MomentumRNN: Integrating Momentum \\ into Recurrent Neural Networks
}

\author{%
  Tan M. Nguyen \\
  Department of ECE\\ 
  Rice University, Houston, USA\\
   \And
   Richard G. Baraniuk\\
  Department of ECE\\ 
  Rice University, Houston, USA\\
   \AND
   Andrea L. Bertozzi \\
  Department of Mathematics\\
  University of California, Los Angeles\\
   \And
   Stanley J. Osher \\
   Department of Mathematics\\
   University of California, Los Angeles\\
   \And
   Bao Wang \thanks{Please correspond to: wangbaonj@gmail.com or mn15@rice.edu} \\
   Department of Mathematics\\
   Scientific Computing and Imaging (SCI) Institute\\
   University of Utah, Salt Lake City, UT, USA\\
}

\begin{document}

\setlength{\abovedisplayskip}{2.5pt}
\setlength{\belowdisplayskip}{2.5pt}

\maketitle

\begin{abstract}
Designing deep neural networks is an art that often involves an expensive search over candidate architectures. To overcome this for recurrent neural nets (RNNs), we establish a connection between the hidden state dynamics in an RNN and gradient descent (GD). We then integrate momentum into this framework and propose a new family of RNNs, called {\em MomentumRNNs}. We theoretically prove and numerically demonstrate that MomentumRNNs alleviate the vanishing gradient issue in training RNNs. We study the momentum long-short term memory (MomentumLSTM) and verify its advantages in convergence speed and accuracy over its LSTM counterpart across a variety of benchmarks. We also demonstrate that MomentumRNN is applicable to many types of recurrent cells, including those in the state-of-the-art orthogonal RNNs. Finally, we show that other advanced momentum-based optimization methods, such as Adam and Nesterov accelerated gradients with a restart, can be easily incorporated into the MomentumRNN framework for designing new recurrent cells with even better performance.
\end{abstract}

\section{Introduction}\label{sec:intro}
Mathematically principled recurrent neural nets (RNNs) facilitate the network design process and reduce the cost of searching over many candidate architectures. A particular advancement in RNNs is the long short-term memory (LSTM) model~\cite{hochreiter1997long} which has achieved state-of-the-art results in many applications, including speech recognition \cite{fernandez2007sequence}, acoustic modeling \cite{sak2014long,qu2017syllable}, and language modeling \cite{palangi2016deep}. There have been many efforts in improving LSTM: \cite{gers1999learning} introduces a forget gate into the original LSTM cell, which can forget information selectively; \cite{gers2000recurrent} further adds peephole connections to the LSTM cell to inspect its current internal states\cite{gers2001lstm}; to reduce the computational cost, a gated recurrent unit (GRU) \cite{cho2014learning} uses a single update gate to replace the forget and input gates in LSTM. Phased LSTM \cite{neil2016phased} adds a new time gate to the LSTM cell and achieves faster convergence than the regular LSTM on learning long sequences. In addition, \cite{rahman2016new} and \cite{pulver2017lstm} introduce a biological cell state and working memory into LSTM, respectively.
Nevertheless, most of RNNs, including LSTMs, 
are biologically informed or even ad-hoc instead of being guided by mathematical principles.

\subsection{Recap on RNNs and LSTM}
Recurrent cells are the building blocks of RNNs. A recurrent cell employs a cyclic connection to update the current hidden state ($\vh_t$) using the past hidden state ($\vh_{t-1}$) and the current input data ($\vx_t$) \cite{elman1990finding}; the dependence of $\vh_t$ on $\vh_{t-1}$ and  $\vx_t$ in a recurrent cell can be written as
\begin{equation}\label{eq:RNN:Cell}
\vh_t = \sigma(\Ub\vh_{t-1} + \Wb\vx_t + \vb),\ \vx_t \in \RR^d,\ \mbox{and}\  \vh_{t-1}, \vh_t \in \RR^h,\ \ t=1, 2, \cdots, T,
\end{equation}
where  $\Ub \in \RR^{h\times h}, \Wb \in \RR^{h\times d}$, and $\vb\in \RR^h$ are trainable parameters; $\sigma(\cdot)$ is a nonlinear activation function, e.g., sigmoid or hyperbolic tangent. Error backpropagation through time is used to train RNN, but it tends to result in exploding or vanishing gradients \cite{bengio1994learning}. Thus RNNs may fail to learn 
long term dependencies. Several approaches exist to improve RNNs' performance, including enforcing unitary weight matrices
\cite{arjovsky2016unitary,wisdom2016full,jing2017tunable,vorontsov2017orthogonality,mhammedi2017efficient,pmlr-v80-helfrich18a}, leveraging LSTM cells, and others \cite{li2018independently,kusupati2018fastgrnn}.

LSTM cells augment the recurrent cell with ``gates'' \cite{hochreiter1997long} and can be formulated as
\begin{equation}\label{eq:LSTM:cell}
\begin{aligned}
\vi_t &= \sigma(\Ub_{ih}\vh_{t-1} + \Wb_{ix}\vx_t + \vb_i),\ \ &(\vi_t: \mbox{input gate})\\
\widetilde{\vc}_t &= \tanh{(\Ub_{\widetilde{c}h}\vh_{t-1} + \Wb_{\widetilde{c}x}\bx_t + \vb_{\widetilde{c}})},\ \ &(\widetilde{\vc}_t: \mbox{cell input})\\
\vc_t &= \vc_{t-1} + \vi_t \odot \widetilde{\vc}_t,\ \ &(\vc_t: \mbox{cell state})\\
\vo_t &= \sigma(\Ub_{oh}\vh_{t-1} + \Wb_{ox}\vx_t + \vb_o ),\ \ &(\vo_t: \mbox{output gate})\\
\vh_t &= \vo_t\odot \tanh{\vc_t},\ \ &(\vh_t: \mbox{hidden state})
\end{aligned}
\end{equation}
where $\Ub_*\in \RR^{h\times h}$, $\Wb_*\in \RR^{h\times d}$, and $\vb_* \in \RR^h$ are learnable parameters, and $\odot$ denotes the Hadamard product. The input gate decides what new information to be stored in the cell state, and the output gate decides what information to output based on the cell state value. The gating mechanism in LSTMs can lead to the issue of saturation \cite{van2018unreasonable,chandar2019towards}.


\subsection{Our Contributions}

\begin{figure}[t]
\centering
\includegraphics[width=1.0\linewidth]{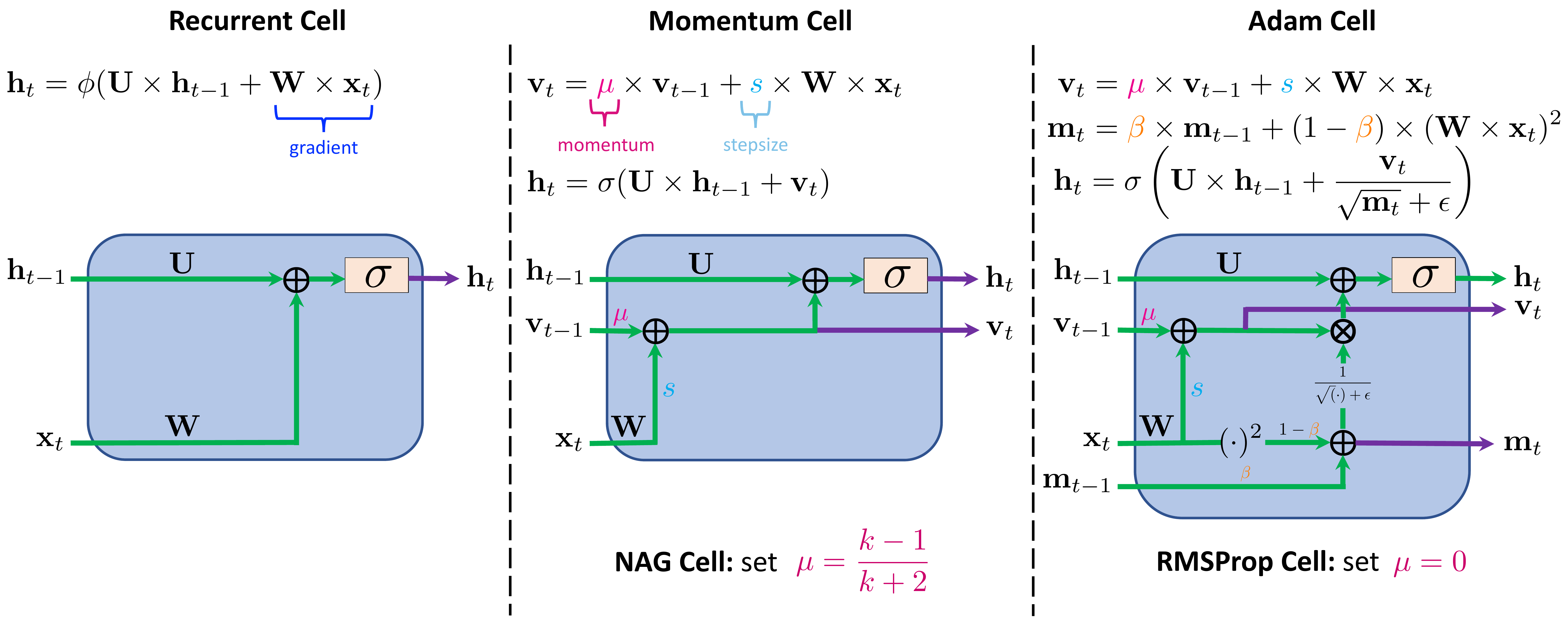}
\vskip -0.35cm
\caption{Illustration of the recurrent cell (left), Momentum/NAG cell (middle), and Adam/RMSProp cell (right). We draw a connection between the dynamics of hidden states in the recurrent cell and GD. We then introduce momentum to recurrent cell as an analogy of the momentum accelerated GD.
}
\label{fig:momentumrnn-vs-rnn}
\end{figure}

In this paper, we develop a gradient descent (GD) analogy of the recurrent cell. In particular, the hidden state update in a recurrent cell is associated with a gradient descent step towards the optimal representation of the hidden state. We then 
propose to integrate momentum that used for accelerating gradient dynamics into the recurrent cell, which results in the momentum cell. At the core of the momentum cell is the use of momentum to accelerate the hidden state learning in RNNs. The architectures of the standard recurrent cell and our momentum cell are illustrated in Fig.~\ref{fig:momentumrnn-vs-rnn}. We provide the design principle and detailed derivation of the momentum cell in Sections~\ref{subsec:momentum-cell} and \ref{sec:Adam:cell}. We call the RNN that consists of momentum cells the MomentumRNN. The major advantages of MomentumRNN are fourfold: 
\begin{itemize}[leftmargin=*]
\item 
MomentumRNN can alleviate the vanishing gradient problem in training RNN.
\item MomentumRNN accelerates training and improves the test accuracy of the baseline RNN. 
\item MomentumRNN is universally applicable to many existing RNNs. It can be easily implemented by changing a few lines of the baseline RNN code.
\item MomentumRNN is principled with theoretical guarantees provided by the momentum-accelerated dynamical system for optimization and sampling. The design principle can be generalized to other advanced momentum-based optimization methods, including Adam~\cite{kingma2014adam} and Nesterov accelerated gradients with a restart~\cite{nesterov1983method,wang2020scheduled}.


\end{itemize}

\subsection{Related Work}
{\bf Dynamical system viewpoint of RNNs.} 
Leveraging the theory of dynamical system to improve RNNs has been an interesting research direction: \cite{laurent2016recurrent} proposes a gated RNN, which is principled by non-chaotical dynamical systems and achieves comparable performance to GRUs and LSTMs. 
\cite{talathi2015improving} proposes a weight initialization strategy inspired by dynamical system theory, which helps the training of RNNs with ReLU nonlinearity. Other RNN algorithms  derived from the dynamical system theories include \cite{
niu2019recurrent,chang2019antisymmetricrnn,chen2019symplectic,kag2019rnns}. Our work is the first that directly integrates momentum into an RNN to accelerate the underlying dynamics and improve the model's performance.



{\bf Momentum in Optimization and Sampling.} Momentum has been a popular technique for accelerating (stochastic) gradient-based optimization \cite{polyak1964some,goh2017momentum,sutskever2013importance,kingma2014adam,bengio2013advances,paszke2019pytorch} and sampling algorithms \cite{duane1987hybrid,neal2011mcmc}
A particularly interesting momentum is the iteration-dependent one in NAG \cite{nesterov1983method,nemirovskii1985optimal,beck2009fast}, which has a significantly better convergence rate than constant momentum for convex optimization. The stochastic gradient NAG that employs a scheduled restart can also be used to accelerate DNN training with better accuracy and faster convergence \cite{wang2020scheduled}.

{\bf Momentum in DNNs.}
Momentum has also been used in designing DNN architectures. \cite{he2019momentum} develops momentum contrast as a way of building large and consistent dictionaries for unsupervised learning with contrastive loss. At the core of this approach is a momentum-based moving average of the queue encoder. Many DNN-based algorithms for sparse coding are designed by unfolding the classical optimization algorithms, e.g., FISTA \cite{beck2009fast}, in which momentum can be used in the underpinning optimizer~\cite{szlam2011structured,chalasani2013fast,mccann2017convolutional,kamilov2016learning,moreau2017understanding}. 
\subsection{Notation}
We denote scalars by lower or upper case letters; vectors and matrices by lower and upper case bold face letters, respectively. For a vector $\vx = (x_1, \cdots, x_d)^T\in \mathbb{R}^d$, we use $\|\vx\| = {(\sum_{i=1}^d |x_i|^2)^{1/2}}$ to denote its $\ell_2$ norm.
For a matrix $\Ab$, we use $\Ab^{\rm T}$ (${\rm T}$ in roman type) and $\Ab^{-1}$ to denote its transpose and inverse, respectively. Also, we denote the spectral norm of $\Ab$ as $\|\Ab\|$. 
We denote the $d$-dimensional standard Gaussian as $\mathcal{N}(\mathbf{0}, \mathbf{I}_{d\times d})$, where $\mathbf{0}$ is the $d$-dimensional zero-vector and $\mathbf{I}_{d\times d}$ is 
an identity matrix. For a function $\phi(\vx): \mathbb{R}^d \rightarrow \mathbb{R}$, we denote $\phi^{-1}(\vx)$ as its inverse and $\nabla \phi(\vx)$ as its gradient.


\section{Momentum RNNs}\label{sec:momentum:nets}
\subsection{Background: Momentum Acceleration for Gradient Based Optimization and Sampling}\label{subsec:review:momentum}

Momentum has been successfully used to accelerate the gradient-based algorithms for optimization and sampling. In optimization, we aim to find a stationary point of a given function $f(\vx), \vx\in \RR^d$. Starting from $\vx_0\in \RR^d$, GD iterates as $\vx_t = \vx_{t-1} - s \nabla f(\vx_t)$ with $s>0$ being the step size. This can be significantly accelerated by using the 
momentum~\cite{sutskever2013importance}, which results in
\begin{equation}
\label{eq:MGD}
\vp_0=\vx_0;\;
\vp_{t} = \mu \vp_{t-1} + s\nabla f(\vx_{t});\;
\vx_{t} = \vx_{t-1} - \vp_{t},\ \ t\geq 1,
\end{equation}
where $\mu \geq 0$ is the momentum constant. In sampling, Langevin Monte Carlo (LMC) \cite{coffey2012langevin} is used to sample from the distribution $\pi \propto \exp\{-f(\vx)\}$, where $\exp\{-f(\vx)\}$ is the probability distribution function. The update at each iteration is given by
\begin{equation}
\label{eq:LMC}
\vx_t = \vx_{t-1} - s\nabla f(\vx_t) + \sqrt{2s}\boldsymbol{\epsilon}_t,
\ s\geq 0,\ t\geq 1,\ \boldsymbol{\epsilon}_t \sim \mathcal{N}(\mathbf{0}, \mathbf{I}_{d\times d}).
\end{equation}
We can also use momentum to accelerate LMC, which results in the following Hamiltonian Monte Carlo (HMC) update \cite{coffey2012langevin}:
\begin{equation}
\label{eq:HMC}
\vp_0=\vx_0;\; \vp_{t} = \vp_{t-1} - \gamma s \vp_{t-1} - s \eta\nabla f(\vx_{t-1}) + \sqrt{2\gamma s\eta} \boldsymbol{\epsilon}_t;\;
\vx_{t} = \vx_{t-1} + s\vp_{t},\ \ t\geq 1,
\end{equation}
where $\boldsymbol{\epsilon}_t \sim \mathcal{N}(\mathbf{0}, \mathbf{I}_{d\times d})$ while $\gamma, \eta, s >0$ are the friction parameter, inverse mass, and step size, resp. 


\subsection{
Gradient Descent Analogy for RNN and MomentumRNN}\label{subsec:momentum-cell}
Now, we are going to establish a connection between RNN and GD, and further leverage momentum to improve RNNs. Let $\widetilde{\Wb} = [\Wb, \vb]$ and $\widetilde{\vx}_t = [\vx_t, 1]^T$ in \eqref{eq:RNN:Cell}, then we have $\vh_t = \sigma(\Ub\vh_{t-1} + \widetilde{\Wb}\widetilde{\vx}_t)$. For the ease of notation, without ambiguity we denote $\Wb := \widetilde{\Wb}$ and $\vx_t := \widetilde{\vx}_t$. Then the recurrent cell can be reformulated as
\begin{equation}
\label{eq:basic:cell}
\vh_t = \sigma(\Ub \vh_{t-1} + \Wb \vx_t).
\end{equation}
Moreover, let $\phi(\cdot) := \sigma(\Ub(\cdot))$ and $\vu_t := \Ub^{-1}\Wb\vx_t$, we can rewrite \eqref{eq:basic:cell} as
\begin{equation}
\label{eq:recurent-cell:PGD}
\vh_t = \phi(\vh_{t-1} + \vu_t).
\end{equation}
If we regard $-\vu_t$ as the ``gradient'' at the $t$-th iteration, then we can consider \eqref{eq:recurent-cell:PGD} as the dynamical system which updates the hidden state by the gradient and then transforms the updated hidden state by the nonlinear activation function $\phi$. We propose the following accelerated dynamical system to accelerate the dynamics of  \eqref{eq:recurent-cell:PGD}, which is principled by the accelerated gradient descent theory (see subsection~\ref{subsec:review:momentum}): 
\begin{align}\label{eq:momentum:PGD}
\vp_{t} = \mu\vp_{t-1} - s \vu_{t};\ \ \vh_{t} =\phi(\vh_{t-1} - \vp_{t}),
\end{align}
where $\mu \geq 0, s >0$ are two hyperparameters, which are the analogies of the momentum coefficient and step size in the momentum-accelerated GD, 
respectively. Let $\bv_t := -\Ub\vp_t$, we arrive at the following dynamical system:
\begin{align}\label{eq:momentum:cell}
\bv_{t} = \mu\bv_{t-1} + s \Wb\vx_{t};\ \ \vh_{t} = \sigma(\Ub\vh_{t-1} + \bv_{t}).
\end{align}
The architecture of the momentum cell that corresponds to the dynamical system \eqref{eq:momentum:cell} is plotted in Fig.~\ref{fig:momentumrnn-vs-rnn} (middle). Compared with the recurrent cell, the momentum cell introduces an auxiliary momentum state in each update and scales the dynamical system with two positive hyperparameters 
$\mu$ and $s$. 

\begin{remark}
\label{rm:different-parameterization}
Different parameterizations of \eqref{eq:momentum:PGD} can result in different momentum cell architectures. For instance, if we let $\bv_t = -\vp_t$, we end up with the following dynamical system:
\begin{align}\label{eq:momentum:cell2}
\bv_{t} = \mu\bv_{t-1} + s \widehat{\Wb}\vx_{t};\ \ \vh_{t} = \sigma(\Ub\vh_{t-1} + \Ub\bv_{t}),
\end{align}
where $\widehat{\Wb} := \Ub^{-1}\Wb$ is the trainable weight matrix. Even though \eqref{eq:momentum:cell} and \eqref{eq:momentum:cell2} are mathematically equivalent, the training procedure might cause the MomentumRNNs that are derived from different parameterizations to have different performances.
\end{remark}

\begin{remark}
We put the nonlinear activation in the second equation of \eqref{eq:momentum:PGD} to ensure that the value of $\vh_t$ is in the same range as the original recurrent cell. 
\end{remark}

\begin{remark}
The derivation above also applies to the dynamical systems in the LSTM cells, and we can design the MomentumLSTM in the same way as designing the MomentumRNN.
\end{remark}













\subsection{Analysis of the Vanishing Gradient Issue: Momentum Cell vs. Recurrent Cell}\label{subsec:gradient}
Let $\vh_T$ and $\vh_t$ be the state vectors at the time step $T$ and $t$, respectively, and we suppose $T\gg t$. Furthermore, assume that $\mathcal{L}$ is the objective to minimize, then
\begin{equation}
\label{eq:gradient:rnn}
{\small \frac{\partial \mathcal L}{\partial \vh_t} = \frac{\partial \mathcal{L}}{\partial\vh_T}\cdot\frac{\partial \vh_T}{\partial \vh_t} = \frac{\partial \mathcal L}{\partial \vh_T}\cdot \prod_{k=t}^{T-1}\frac{\partial \vh_{k+1}}{\partial \vh_k} = \frac{\partial \mathcal L}{\partial \vh_T}\cdot \prod_{k=t}^{T-1}(\Db_k\Ub^{\rm T}),}
\end{equation}
where {\small $\Ub^{\rm T}$} is the transpose of $\Ub$ and {\small $\Db_k = {\rm diag}(\sigma'(\Ub\vh_k + \Wb\vx_{k+1}))$} is a diagonal matrix with $\sigma'(\Ub\vh_k + \Wb\vx_{k+1})$ being its diagonal entries. 
{\small $\|\prod_{k=t}^{T-1}(\Db_k\Ub^{\rm T})\|_2$} tends to either vanish or explode \cite{bengio1994learning}. We can use regularization or gradient clipping to mitigate the exploding gradient, leaving vanishing gradient as the major obstacle to training RNN to learn long-term dependency~\cite{pascanu2013difficulty}.  
We can rewrite \eqref{eq:momentum:cell} as
\vskip -0.5cm
\begin{equation}
\label{eq:momentum:cell:one:eq}
{\small \vh_t = \sigma\left(\Ub(\vh_{t-1} - \mu\vh_{t-2}) + \mu\sigma^{-1}(\vh_{t-1}) + s\Wb\vx_t\right),}
\end{equation}
where $\sigma^{-1}(\cdot)$ is the inverse function of $\sigma(\cdot)$. We compute $\partial \mathcal L/\partial \vh_t$ as follows
\begin{equation}
\label{eq:gradient:momentum}
{\small \frac{\partial \mathcal L}{\partial \vh_t} = \frac{\partial \mathcal{L}}{\partial\vh_T}\cdot\frac{\partial \vh_T}{\partial \vh_t} = \frac{\partial \mathcal L}{\partial \vh_T}\cdot \prod_{k=t}^{T-1}\frac{\partial \vh_{k+1}}{\partial \vh_k} = \frac{\partial \mathcal L}{\partial \vh_T}\cdot \prod_{k=t}^{T-1}\widehat{\Db}_k[\Ub^{\rm T} + \mu\boldsymbol\Sigma_k],}
\end{equation}
where {\small $\widehat{\Db}_k = {\rm diag}(\sigma'(\Ub(\vh_{k} - \mu\vh_{k-1}) + \mu\sigma^{-1}(\vh_{k}) + s\Wb\vx_{k+1} ))$} and {\small $\boldsymbol\Sigma = {\rm diag}((\sigma^{-1})'(\vh_k))$}. For mostly used $\sigma$, e.g., sigmoid and tanh,  $(\sigma^{-1}(\cdot))' > 1$ 
and $\mu\boldsymbol\Sigma_k$ dominates $\Ub^{\rm T}$.\footnote{In the vanishing gradient scenario, $\|\Ub\|_2$ is small; also it can be controlled by regularizing the loss function.} Therefore, with an appropriate choice of $\mu$, the momentum cell can alleviate vanishing gradient and accelerate training. 

\begin{figure}[t!]
\centering
\includegraphics[width=0.99\linewidth]{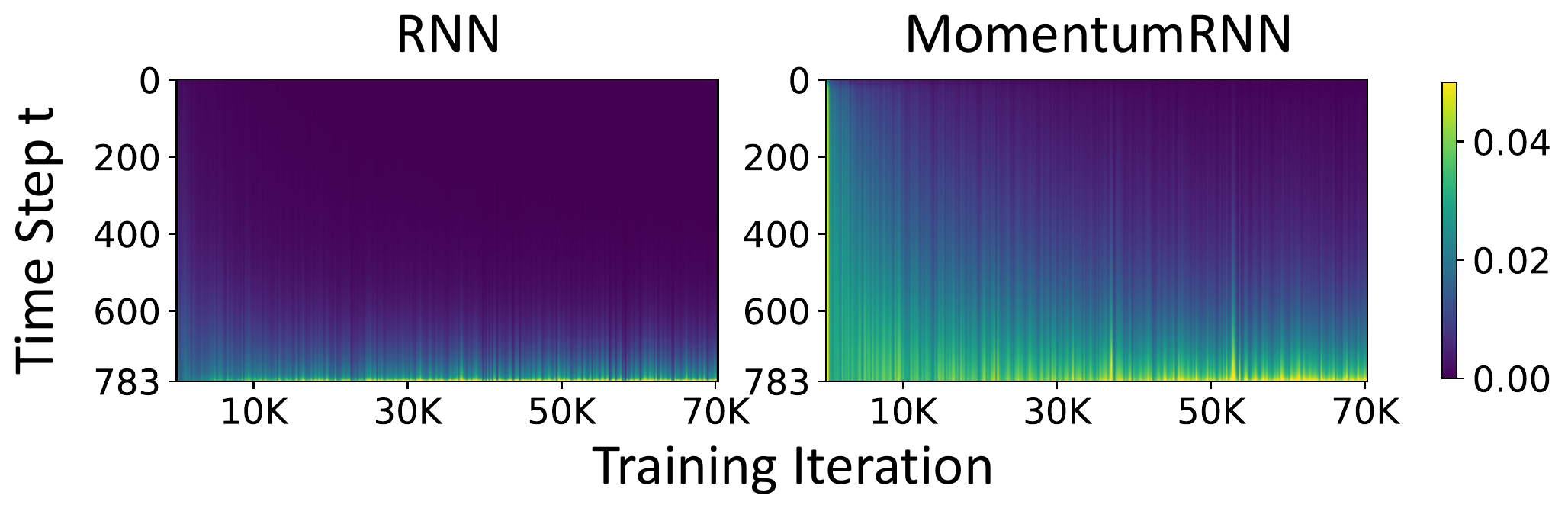}
\vskip -0.2cm
\caption{$\ell_2$ norm of the gradients of the loss $\mathcal{L}$ w.r.t. the state vector $\vh_t$ at each time step $t$ for RNN (left) and MomentumRNN (right). MomentumRNN does not suffer from vanishing gradients.}
\label{fig:stability}
\end{figure}

We empirically corroborate that momentum cells can alleviate vanishing gradients by training a MomentumRNN and its corresponding RNN on the PMNIST classification task and plot $\|\partial \mathcal L/\partial \vh_t\|_{2}$ for each time step $t$. Figure~\ref{fig:stability} confirms that unlike in RNN, the gradients in MomentumRNN do not vanish. More details on this experiment are provided in the Appendix~\ref{sec:exp:details}.

\subsection{Beyond MomentumRNN: NAG and Adam Principled Recurrent Neural Nets}\label{sec:Adam:cell} 
There are several other advanced formalisms of momentum existing in optimization, which can be leveraged for RNN architecture design. In this subsection, we present two additional variants of MomentumRNN that are derived from the Nesterov accelerated gradient (NAG)-style momentum with restart \cite{nesterov1983method,wang2020scheduled} and Adam \cite{kingma2014adam}.

{\bf NAG Principled RNNs.} The momentum-accelerated GD can be further accelerated by  replacing the constant momentum coefficient $\mu$ in~\eqref{eq:momentum:cell} with the NAG-style momentum, i.e. setting $\mu$ to $(t-1)/(t+2)$ at the $t$-th iteration. Furthermore, we can accelerate NAG by resetting the momentum to 0 after every $F$ iterations, i.e. $\mu = (t \mod F)/((t\mod F) + 3)$, which is the NAG-style momentum with a scheduled restart of the appropriately selected frequency $F$ \cite{wang2020scheduled}. For convex optimization, NAG has a convergence rate $O(1/t^2)$, which is significantly faster than GD or GD with constant momentum whose convergence rate is $O(1/t)$. Scheduled restart not only accelerates NAG to a linear convergence rate $O(\alpha^{-t}) (0<\alpha <1)$ under mild extra assumptions but also stabilizes the NAG iteration \cite{wang2020scheduled}.
We call the MomentumRNN with the NAG-style momentum and scheduled restart momentum the NAG-based RNN and the scheduled restart RNN (SRRNN), respectively.

{\bf Adam Principled RNNs.} Adam~\cite{kingma2014adam} leverages the moving average of historical gradients and entry-wise squared gradients to accelerate the stochastic gradient dynamics. We use Adam to accelerate \eqref{eq:recurent-cell:PGD} and end up with the following iteration
\begin{equation}
\label{eq:Adam:cell}
{\small \vp_t = \mu \vp_{t-1} + (1-\mu)\vu_t;\ \vm_t = \beta\vm_{t-1} + (1-\beta)\vu_t\odot\vu_t;\ \vh_t = \phi(\vh_{t-1} - s \frac{\vp_t}{\sqrt{\vr_t} +\epsilon}),}
\end{equation}
where $\mu, s, \beta > 0$ are hyperparameters, $\epsilon$ is a small constant and chosen to be $10^{-8}$ by default, and $\odot$/$\sqrt{\cdot}$ denotes the entrywise product/square root\footnote{In contrast to Adam, we do not normalize $\vp_t$ and $\vm_t$ since they can be absorbed in the weight matrices.}. Again, let $\bv_t = -\Ub\vp_t$, we rewrite \eqref{eq:Adam:cell} as follows
\begin{equation}
\label{eq:Adam:momentum:cell}
\hskip -0.3cm{\footnotesize \bv_t = \mu \bv_{t-1} + (1-\mu)\Wb\vx_t;\ \vm_t = \beta\vm_{t-1} + (1-\beta)\vu_t\odot\vu_t;\ \vh_t = \sigma(\Ub\vh_{t-1} + s \frac{\bv_t}{\sqrt{\vm_t} +\epsilon}).} \nonumber
\end{equation}
As before, here $\vu_t := \Ub^{-1}\Wb\vx_t$. Computing $\Ub^{-1}$ is expensive. Our experiments suggest that replacing $\vu_t\odot\vu_t$ by $\Wb\vx_t \odot \Wb\vx_t$ is sufficient and more efficient to compute. In our implementation, we also 
relax $\bv_t = \mu \bv_{t-1} + (1-\mu)\Wb\vx_t$ 
to $\bv_t = \mu \bv_{t-1} + s\Wb\vx_t$  that follows the momentum in the MomentumRNN \eqref{eq:momentum:cell}  
for better performance. Therefore, we propose the \emph{AdamRNN} that is given by
\begin{equation}
\label{eq:Adam:momentum:cell:final}
{\small \bv_t = \mu \bv_{t-1} + s\Wb\vx_t;\ \ \vm_t = \beta\vm_{t-1} + (1-\beta)(\Wb\vx_t\odot\Wb\vx_t);\ \ \vh_t = \sigma(\Ub\vh_{t-1} + \frac{\bv_t}{\sqrt{\vm_t} +\epsilon}).}
\end{equation}
In AdamRNN, if $\mu$ is set to 0, we achieve another new RNN, which obeys the RMSProp gradient update rule~\cite{Tieleman2012}. We call this new model the \emph{RMSPropRNN}.
\begin{remark}
Both AdamRNN and RMSPropRNN can also be derived by letting $\bv_t = -\vp_t$ and $\widehat{\Wb} := \Ub^{-1}\Wb$ as in Remark~\ref{rm:different-parameterization}. This parameterization yields the following formulation for AdamRNN
\begin{equation}
\label{eq:Adam:momentum:cell:final-2}
{\small \bv_t = \mu \bv_{t-1} + s\widehat{\Wb}\vx_t;\ \ \vm_t = \beta\vm_{t-1} + (1-\beta)(\widehat{\Wb}\vx_t\odot\widehat{\Wb}\vx_t);\ \ \vh_t = \sigma(\Ub\vh_{t-1} + \frac{\Ub\bv_t}{\sqrt{\vm_t} +\epsilon}).} \nonumber
\end{equation}
Here, we simply need to learn 
$\widehat{\Wb}$ and $\Ub$ without any relaxation. In contrast, we relaxed $\Ub^{-1}$ to an identity matrix in~\eqref{eq:Adam:momentum:cell:final}. Our experiments suggest that both parameterizations
yield similar results.
\end{remark}

\section{Experimental Results}\label{sec:experiments}
In this section, we evaluate the effectiveness of our momentum approach in designing RNNs in terms of convergence speed and accuracy. We compare the performance of the MomentumLSTM with the baseline LSTM~\cite{hochreiter1997long} in the following tasks: 1) the object classification task on pixel-permuted MNIST~\cite{le2015simple}, 2) the speech prediction task on the TIMIT dataset~\cite{arjovsky2016unitary, pmlr-v80-helfrich18a, wisdom2016full, mhammedi2017efficient, pmlr-v48-henaff16}, 3) the celebrated copying and adding tasks \cite{hochreiter1997long, arjovsky2016unitary}, and 4) the language modeling task on the Penn TreeBank (PTB) dataset~\cite{mikolov2010recurrent}. These four tasks are among standard benchmarks to measure the performance of RNNs and their ability to handle long-term dependencies. Also, these tasks cover different data modalities -- image, speech, and text data -- as well as a variety of model sizes, ranging from thousands to millions of parameters with one (MNIST and TIMIT tasks) or multiple (PTB task) recurrent cells in concatenation. Our experimental results confirm that MomentumLSTM converges faster and yields better test accuracy than the baseline LSTM across tasks and settings. We also discuss the AdamLSTM, RMSPropLSTM, and scheduled restart LSTM (SRLSTM) and show their advantage over MomentumLSTM in specific tasks. Computation time and memory cost of our models versus the baseline LSTM are provided in Appendix~\ref{sec:appendix:Time-and-Memory}. All of our results are averaged over 5 runs with different seeds.
%
We include details on the models, datasets, training procedure, and hyperparameters used in our experiments in Appendix~\ref{sec:exp:details}. For MNIST and TIMIT experiments, we use the baseline codebase provided by~\cite{dtrivgithub}. For PTB experiments, we use the baseline codebase provided by~\cite{ptbgithub}.

\subsection{Pixel-by-Pixel MNIST}
\label{sec:mnist-exp}
In this task, we classify image samples of hand-written digits from the MNIST dataset~\cite{lecun2010mnist} into one of the ten classes. Following the implementation of~\cite{le2015simple}, we flatten the image of original size 28 $\times$ 28 pixels and feed it into the model as a sequence of length 784. In the unpermuted task (MNIST), the sequence of pixels is processed row-by-row. In the permuted task (PMNIST), a fixed permutation is selected at the beginning of the experiments and then applied to both training and test sequences. We summarize the results in Table~\ref{tab:mnist}. Our experiments show that \emph{MomentumLSTM achieves better test accuracy than the baseline LSTM in both MNIST and PMNIST digit classification tasks} using different numbers of hidden units (i.e. $N=128, 256$). Especially, the improvement is significant on the PMNIST task, which is designed to test the performance of RNNs in the context of long-term memory. Furthermore, we notice that \emph{MomentumLSTM converges faster than LSTM} in all settings. Figure~\ref{fig:loss-vs-iters} (left two panels) corroborates this observation when using $N=256$ hidden units.

\begin{table}[!t]
\vspace{-0.07in}
    \caption{Best test accuracy at the MNIST and PMNIST tasks (\%). 
    We use the baseline results reported in \cite{pmlr-v80-helfrich18a},~\cite{wisdom2016full},~\cite{vorontsov2017orthogonality}. All of our proposed models outperform the baseline LSTM. Among the models using $N=256$ hidden units, RMSPropLSTM yields the best results in both tasks.}
\vspace{-0.1in}
\label{tab:mnist}
\begin{center}
\begin{footnotesize}
\begin{sc}
\begin{tabular}{lclcc}
    \toprule
    Model & n & \# params & MNIST & PMNIST \\
    \midrule
    \midrule
    LSTM & $128$ & $\approx 68K$ & $98.70$\cite{pmlr-v80-helfrich18a},$97.30$  \cite{vorontsov2017orthogonality} & $92.00$  \cite{pmlr-v80-helfrich18a},$92.62$ \cite{vorontsov2017orthogonality} \\
    LSTM & $256$ & $\approx 270K$ & $98.90$ \cite{pmlr-v80-helfrich18a}, $98.50$ \cite{wisdom2016full} & $92.29$  \cite{pmlr-v80-helfrich18a}, $92.10$  \cite{wisdom2016full} \\
    \midrule
    MomentumLSTM & $128$ & $\approx 68K$ & $\bf{99.04 \pm 0.04}$ & $\bf{93.40 \pm 0.25}$\\
    MomentumLSTM & $256$ & $\approx 270K$ & $\bf{99.08 \pm 0.05}$ & $\bf{94.72 \pm 0.16}$\\
    \midrule
    AdamLSTM & $256$ & $\approx 270K$ & $99.09 \pm 0.03$ & $95.05 \pm 0.37$\\
    RMSPropLSTM & $256$ & $\approx 270K$ & $\bf{99.15 \pm 0.06}$ & $\bf{95.38 \pm 0.19}$\\
    SRLSTM & $256$ & $\approx 270K$ & $ 99.01 \pm 0.07 $ & $93.82 \pm 1.85$\\
    \bottomrule
\end{tabular}
\end{sc}
\end{footnotesize}
\end{center}
\end{table}

\subsection{TIMIT Speech Dataset}
\label{sec:timit-exp}
We study how MomentumLSTM performs on audio data with speech prediction experiments on the TIMIT speech dataset~\cite{garofolo1993timit}, which is a collection of real-world speech recordings. As first proposed by~\cite{wisdom2016full}, the recordings are downsampled to 8kHz and then transformed into log-magnitudes via a short-time Fourier transform (STFT). The task accounts for predicting the next log-magnitude given the previous ones. We use the standard train/validation/test separation in~\cite{wisdom2016full, lezcano2019cheap, casado2019trivializations}, thereby having 3640 utterances for the training set with a validation set of size 192 and a test set of size 400.

\begin{figure}[!t]
\centering
\includegraphics[width=1.0\linewidth]{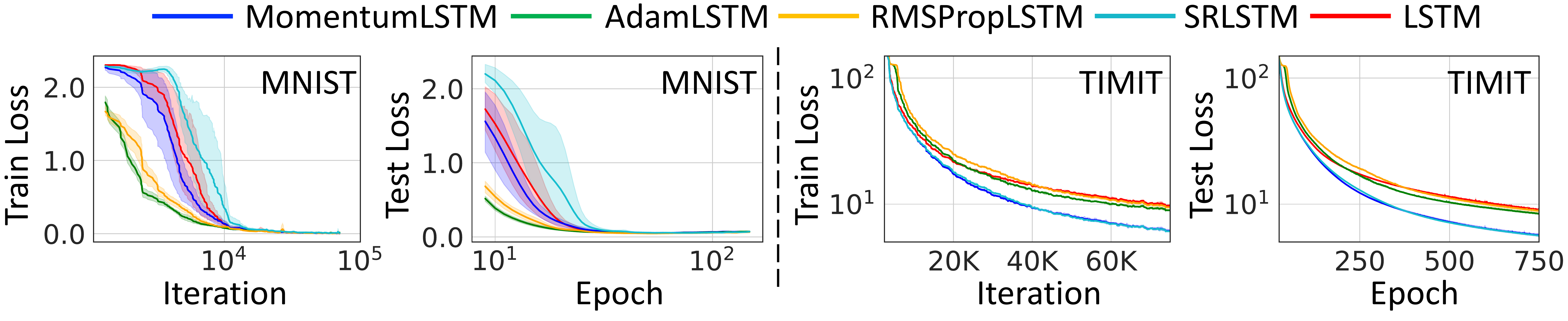}
\vskip -0.2cm
\caption{Train and test loss  of MomentumLSTM (blue), AdamLSTM (green), RMSPropLSTM (orange), SRLSTM (cyan), and LSTM (red) for MNIST (left two panels) and TIMIT (right two panels) tasks. MomentumLSTM converges faster than LSTM in both tasks. For MNIST, AdamLSTM and RMSPropLSTM converge fastest. For TIMIT, MomentumLSTM and SRLSTM converge fastest.}
\label{fig:loss-vs-iters}
\end{figure}

The results for this TIMIT speech prediction are shown in Table~\ref{tab:timit}. Results are reported on the test set using the model parameters
that yield the best validation loss. Again, we see the advantage of MomentumLSTM over the baseline LSTM. In particular, MomentumLSTM yields much better prediction accuracy and faster convergence speed compared to
LSTM. Figure~\ref{fig:loss-vs-iters} (right two panels)
shows the convergence of MomentumLSTM vs. LSTM when using $N=158$ hidden units.

{\bf Remark:} The TIMIT dataset is not open for
public, so we do not
have access to the preprocessed 
data from previous papers. We followed the data preprocessing in~\cite{wisdom2016full, lezcano2019cheap, casado2019trivializations} to generate the preprocessed data for our experiments and did our best to reproduce the baseline results.
In Table~\ref{tab:timit} and~\ref{tab:timit-dtriv}, we include both our reproduced results and the ones reported from previous works.

\begin{table}[!t]
\vspace{-0.1in}
    \caption{Test and validation MSEs at the end of the epoch with the lowest validation
MSE for the TIMIT task. All of our proposed models outperform the baseline LSTM. Among models using $N=158$ hidden units, SRLSTM performs the best.} 
\vspace{-0.2in}
\label{tab:timit}
\begin{center}
\begin{footnotesize}
\begin{sc}
\begin{tabular}{lclcc}
    \toprule
    Model & n & \# params & Val. MSE & Test MSE \\
    \midrule
    \midrule
    LSTM & $84$ & $\approx 83K$ & $14.87 \pm 0.15$ ($15.42$ \cite{pmlr-v80-helfrich18a,lezcano2019cheap}) & $14.94 \pm 0.15$ ($14.30$ \cite{pmlr-v80-helfrich18a,lezcano2019cheap}) \\
    LSTM & $120$ & $\approx 135K$ & $11.77 \pm 0.14$ ($13.93$ \cite{pmlr-v80-helfrich18a,lezcano2019cheap}) & $11.83 \pm 0.12$ ($12.95$ \cite{pmlr-v80-helfrich18a,lezcano2019cheap}) \\
    LSTM & $158$ & $\approx 200K$ & $9.33 \pm 0.14$ ($13.66$ \cite{pmlr-v80-helfrich18a,lezcano2019cheap}) & $9.37 \pm 0.14$ ($12.62$ \cite{pmlr-v80-helfrich18a,lezcano2019cheap}) \\
    \midrule
    MomentumLSTM & $84$ & $\approx 83K$ & $\bf{10.90 \pm 0.19}$ & $\bf{10.98 \pm 0.18}$\\
    MomentumLSTM & $120$ & $\approx 135K$ & $\bf{8.00 \pm 0.30}$ & $\bf{8.04 \pm 0.30}$\\
    MomentumLSTM & $158$ & $\approx 200K$ & $\bf{5.86 \pm 0.14}$ & $\bf{5.87 \pm 0.15}$\\
    \midrule
    AdamLSTM & $158$ & $\approx 200K$ & $8.66 \pm 0.15$ & $8.69 \pm 0.14$\\
    RMSPropLSTM & $158$ & $\approx 200K$ & $9.13 \pm 0.33$ & $9.17 \pm 0.33$\\
    SRLSTM & $158$ & $\approx 200K$ & $\bf{5.81 \pm 0.10}$ & $\bf{5.83 \pm 0.10}$\\
    \bottomrule
\end{tabular}
\end{sc}
\end{footnotesize}
\end{center}
\vskip -0.05in
\end{table}


\subsection{Copying and Adding Tasks}
\label{sec:copying-adding-maintext}
\begin{figure}[t!]
\centering
\includegraphics[width=0.75\linewidth]{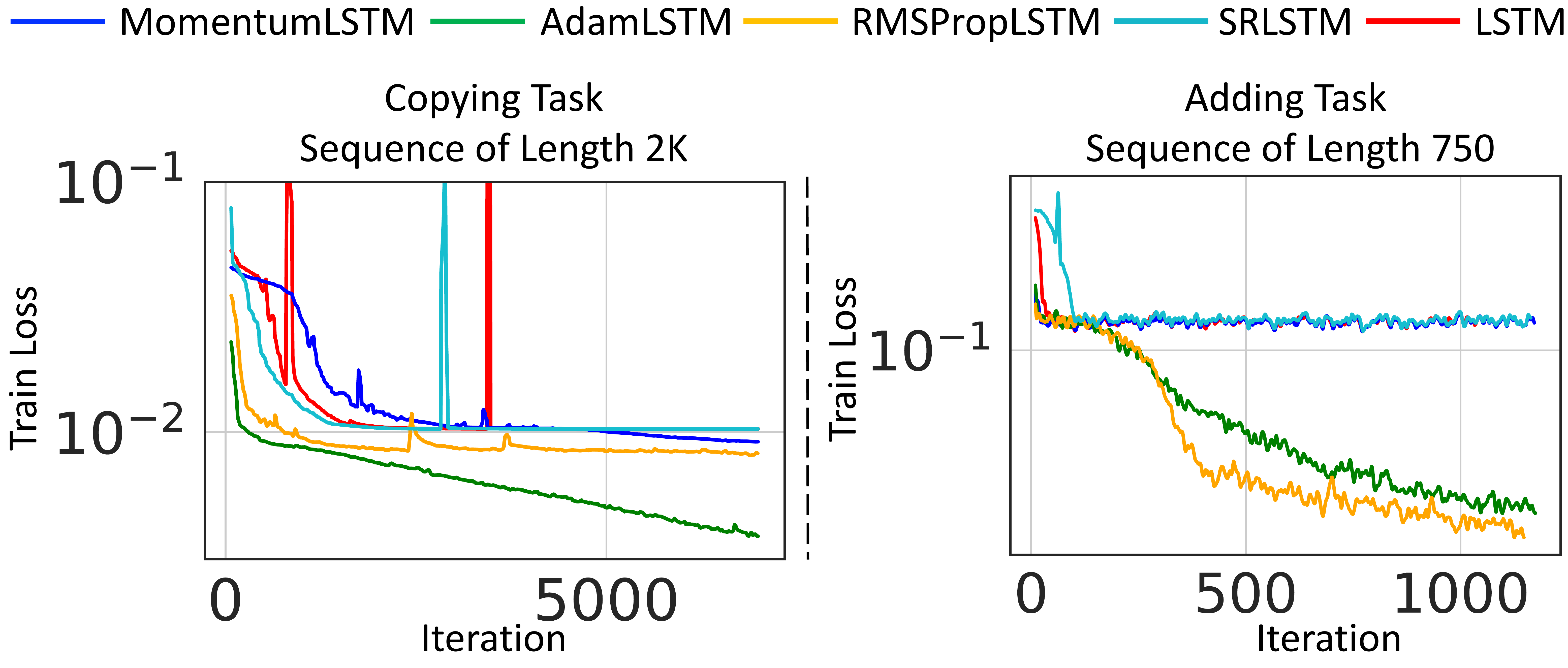}
\caption{Train loss vs. iteration for (left) copying task with sequence length 2K and (right) adding task with sequence length 750. AdamLSTM and RMSPropLSTM converge faster and to better final losses than other models. MomentumLSTM and SRLSTM converge to similar losses as LSTM.}
\label{fig:copy-add-task}
\end{figure}

Two other important tasks for measuring the ability of a model to learn long-term dependency are the copying and adding tasks \cite{hochreiter1997long, arjovsky2016unitary}. In both copying and adding tasks, avoiding vanishing/exploding gradients becomes more relevant when the input sequence length increases. We compare the performance of MomentumLSTM over LSTM on these tasks. We also examine the performance of AdamLSTM, RMSPropLSTM, and SRLSTM on the same tasks. We define the copying and adding tasks in Appendix~\ref{sec:copying-adding-appendix} and summarize our results in Figure~\ref{fig:copy-add-task}. In copying task for sequences of length 2K, MomentumLSTM obtains slightly better final training loss than the baseline LSTM (0.009 vs.\ 0.01). In adding task for sequence of length 750, both models achieve similar training loss of 0.162. However, AdamLSTM and RMSPropLSTM significantly outperform the baseline LSTM.

\subsection{Word-Level Penn TreeBank}
To study the advantage of MomentumLSTM over LSTM on text data, we perform language
modeling on a preprocessed version of the  
PTB dataset~\cite{mikolov2010recurrent}, 
which has been a standard benchmark for evaluating language models. 
Unlike the baselines used in 
\begin{wrapfigure}[12]{r}{.5\textwidth}
\vspace{-0.15 in}
\includegraphics[width=\linewidth]{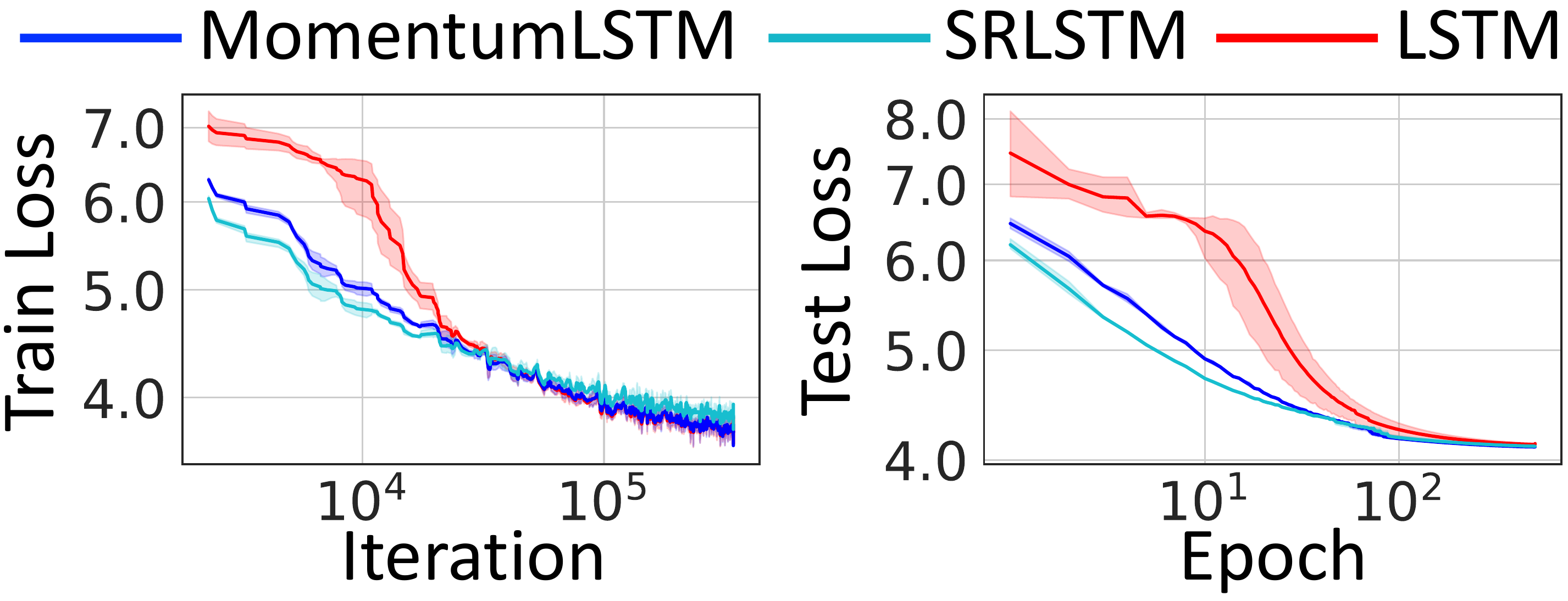}
\vspace{-0.2 in}
\caption{Train (left) and test loss (right) of MomentumLSTM (blue), SRLSTM (cyan), and LSTM (red) for the Penn Treebank language modeling tasks at word level.}\label{fig:train-loss-vs-iters-ptb}
\end{wrapfigure}
the (P)MNIST and TIMIT experiments which 
contain one LSTM cell, in this PTB experiment, we use a three-layer LSTM model, 
which contains three concatenated LSTM cells, as the baseline. The size of this model in terms of the number of parameters is also much larger than those in the (P)MNIST and TIMIT experiments. Table~\ref{tab:ptb-word-level} shows the test and validation perplexity (PPL) using the model parameters that yield the best validation loss. Again, MomentumLSTM achieves better perplexities and converges faster than the baseline LSTM (see Figure~\ref{fig:train-loss-vs-iters-ptb}).
\begin{table}[t!]
\vskip -0.2in
    \caption{Model test perplexity at the end of the epoch with the lowest validation perplexity for the Penn Treebank language modeling task (word level).} 
\vspace{-0.1in}
\label{tab:ptb-word-level}
\begin{center}
\begin{small}
\begin{sc}
\begin{tabular}{l l c c}
    \toprule
    Model & \# params & Val. PPL & Test PPL \\
    \midrule
    \midrule
    \lstm & $\approx 24M$ & $61.96 \pm 0.83$ & $59.71 \pm 0.99$ ($58.80$ \cite{merity2018regularizing}) \\
    \midrule
    MomentumLSTM & $\approx 24M$ & $\bf{60.71 \pm 0.24}$ & $\bf{58.62 \pm 0.22}$\\
    \midrule
    SRLSTM & $\approx 24M$ & $61.12 \pm 0.68$ & $58.83 \pm 0.62$\\
    \bottomrule
\end{tabular}
\end{sc}
\end{small}
\end{center}
\vskip -0.05in
\end{table}


\subsection{NAG and Adam Principled Recurrent Neural Nets}

We evaluate AdamLSTM, RMSPropLSTM and SRLSTM on all tasks. 
For (P)MNIST and TIMIT tasks, we summarize the test accuracy of the trained models in Tables~\ref{tab:mnist} and~\ref{tab:timit} and provide the plots of train and test losses in Figure~\ref{fig:loss-vs-iters}. We observe that though AdamLSTM and RMSPropLSTM work better than the MomentumLSTM at (P)MNIST task, they yield worse results at the TIMIT task. Interestingly, SRLSTM shows an opposite behavior - better than MomentunLSTM at TIMIT task but worse at (P)MNIST task. For the copying and adding tasks, Figure~\ref{fig:copy-add-task} shows that AdamLSTM and RMSPropLSTM converge faster and to better final training loss than other models in both tasks. Finally, for the PTB task, both MomentumLSTM and SRLSTM outperform the baseline LSTM (see Figure~\ref{fig:train-loss-vs-iters-ptb} and Table~\ref{tab:ptb-word-level}). However, in this task, AdamLSTM and RMSPropLSTM yields slightly worse performance than the baseline LSTM. In particular, test PPL for AdamLSTM and RMSPropLSTM are $61.11 \pm 0.31$, and $64.53 \pm 0.20$, respectively, which are higher than the test PPL for LSTM ($59.71 \pm 0.99$). We observe that there is no model that win in all tasks. This is somewhat expected, given the connection between our model and its analogy to optimization algorithm. An optimizer needs to be chosen for each particular task, and so is for our MomentumRNN. All of our models outperform the baseline LSTM.


\section{Additional Results and Analysis}\label{sec:empirical-analysis}
\begin{figure}[!t]
\centering
\includegraphics[width=1.0\linewidth]{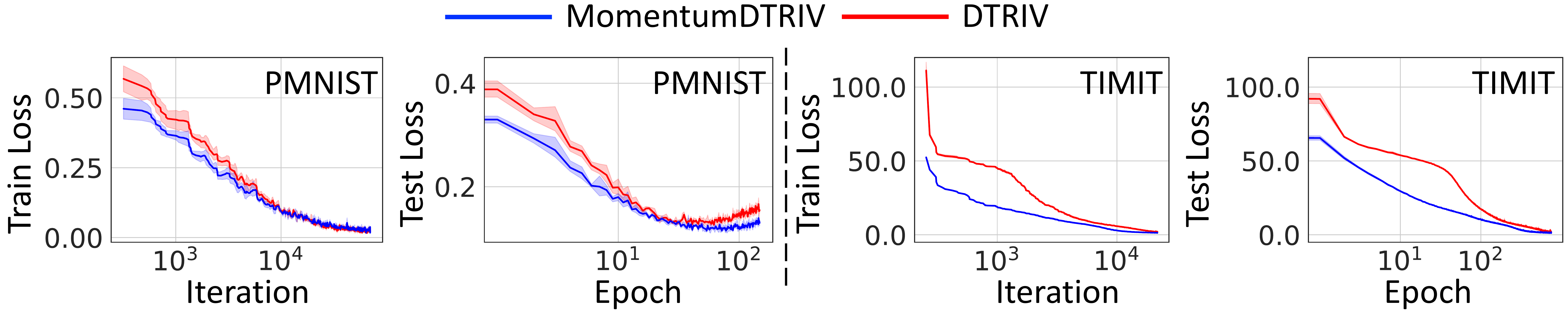}
\vskip -0.1cm
\caption{Train and test loss of MomentumDTRIV (blue) and DTRIV (red) for PMNIST (left two panels) and TIMIT (right two panels) tasks. MomentumDTRIV converges faster than DTRIV in both tasks. For PMNIST task, DTRIV suffers from overtting while MomentumDTRIV overfits less.}
\label{fig:loss-vs-iters-dtriv}
\end{figure}

{\bf Beyond LSTM.}
\label{sec:beyond-lstm}
Our interpretation of hidden state dynamics in RNNs as GD steps and the use of momentum to accelerate the convergence speed and improve the generalization of the model apply to many types of RNNs but not only LSTM. We show the applicability of our momentum-based design approach beyond LSTM by performing PMNIST and TIMIT experiments using the orthogonal RNN equipped with dynamic trivialization (DTRIV)~\cite{casado2019trivializations}. DTRIV is currently among state-of-the-art models for PMNIST digit classification and TIMIT speech prediction tasks. 
Tables~\ref{tab:mnist-dtriv} and~\ref{tab:timit-dtriv} consist of results for our method, namely MomentumDTRIV, in comparison with the baseline results. Again, MomentumDTRIV outperforms the baseline DTRIV by a margin in both PMNIST and TIMIT tasks while converging faster and overfitting less (see Figure~\ref{fig:loss-vs-iters-dtriv}). Results for AdamDTRIV, RMSPropDTRIV, and SRDTRIV on the PMNIST task are provided in Appendix~\ref{sec:appendix:more:exp:results}.

\begin{table}[!t]
\vspace{-0.05in}
    \caption{Best test accuracy on the PMNIST tasks (\%) for MomentumDTRIV and DTRIV. We provide both our reproduced baseline results and those reported in~\cite{casado2019trivializations}. MomentumDTRIV yields better results than the baseline DTRIV in all settings.}
\vspace{-0.1in}
\label{tab:mnist-dtriv}
\begin{center}
\begin{small}
\begin{sc}
\begin{tabular}{clcc}
    \toprule
     n & \# params & PMNIST (DTRIV) & PMNIST (MomentumDTRIV) \\
    \midrule
    \midrule
     $170$ & $\approx 16K$  & $95.21 \pm 0.10$ ($95.20$ \cite{casado2019trivializations}) & $\bf{95.37 \pm 0.09}$\\
     $360$ & $\approx 69K$  & $96.45 \pm 0.10$ ($96.50$ \cite{casado2019trivializations}) & $\bf{96.73 \pm 0.08}$\\
     $512$ & $\approx 137K$  & $96.62 \pm 0.12$ ($96.80$ \cite{casado2019trivializations}) & $\bf{96.89 \pm 0.08}$\\
    \bottomrule
\end{tabular}
\end{sc}
\end{small}
\end{center}
\end{table}

\begin{table}[!t]
    \caption{Test and validation MSE of MomentumDTRIV vs. DTRIV at the epoch with the lowest validation MSE for the TIMIT task. MomentumDTRIV yields much better results than DTRIV.}
\vspace{-0.1in}
\label{tab:timit-dtriv}
\begin{center}
\begin{small}
\begin{sc}
\begin{tabular}{lclcc}
    \toprule
    Model & n & \# params & Val. MSE & Test MSE \\
    \midrule
    DTRIV & $224$ & $\approx 83K$ & $4.74 \pm 0.06$ ($4.75$ \cite{casado2019trivializations}) & $4.70 \pm 0.07$ ($4.71$ \cite{casado2019trivializations}) \\
    DTRIV & $322$ & $\approx 135K$ & $1.92 \pm 0.17$ ($3.39$ \cite{casado2019trivializations}) & $1.87 \pm 0.17$ ($3.76$ \cite{casado2019trivializations}) \\
    \midrule
    MomentumDTRIV & $224$ & $\approx 83K$ & $\bf{3.10 \pm 0.09}$ & $\bf{3.06 \pm 0.09}$ \\
    MomentumDTRIV & $322$ & $\approx 135K$ & $\bf{1.21 \pm 0.05}$ & $\bf{1.17 \pm 0.05}$ \\
    \bottomrule
\end{tabular}
\end{sc}
\end{small}
\end{center}
\end{table}

{\bf Computational Time Comparison.} We study the computational efficiency of the proposed momentum-based models by comparing the time for our models to reach the same test accuracy for LSTM. When training on the PMNIST task using 256 hidden units, we observe that to reach 92.29\% test accuracy for LSTM, LSTM needs 767 min while MomentumLSTM, AdamLSTM, RMSPropLSTM, and SRLSTM only need 551
min, $\bf 225 min$, 416 min, and 348 min, respectively. More detailed results are provided in Appendix~\ref{sec:appendix:Time-and-Memory}.

\begin{figure}[!t]
\centering
\vskip -0.2cm
\includegraphics[width=1.0\linewidth]{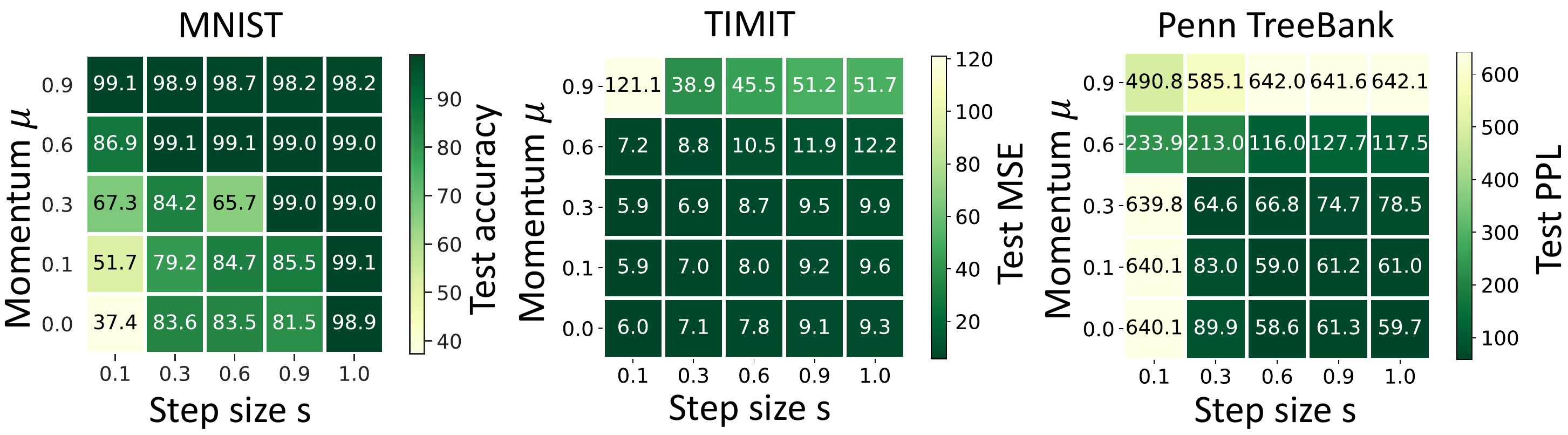}
\vskip -0.2cm
\caption{Ablation study of the effects of momentum and step size on MomentumLSTM's performance.  
We use $N=256$/$158$ hidden units for MNIST/TIMIT task. 
Green denotes better results.
}
\label{fig:ablation-stepsize-mu}
\end{figure}

{\bf Effects of Momentum and Step Size.} To better understand the effects of momentum and step size on the final performance of the trained MomentumLSTM models, we do an ablation study and include the results in Figure~\ref{fig:ablation-stepsize-mu}. The result in each cell is averaged over 5 runs.



\section{Conclusion}\label{sec:conclusion}
In this paper, we propose a universal framework for integrating momentum into RNNs. The resulting MomentumRNN achieves significant acceleration in training and remarkably better performance on the benchmark sequential data prediction tasks over the RNN counterpart. From a theoretical viewpoint, it would be interesting to derive a theory to decipher why training MomentumRNN converges faster and generalizes better. From the neural architecture design perspective, it would be interesting to integrate momentum into the design of the standard convolutional and graph convolutional neural nets. Moreover, the current MomentumRNN requires calibration of the momentum and step size-related hyperparameters;  developing an adaptive momentum for MomentumRNN is of interest.




\clearpage
\section{Broader Impact and Ethical Considerations}
Recurrent neural net (RNN) is among the most important classes of deep learning models. Improving training efficiency and generalization performance of RNNs not only advances image classification and language modeling but also benefits epidemiological models for pandemic disease prediction. RNNs have also been successfully used for the molecular generation \cite{kotsias2019direct}. Developing better RNNs that enable modeling of long term dependency, such as our Momentum RNN,  has the potential to facilitate life science research.  
In order to fullfill that potential, more development is needed. For example, the current MomentumRNN requires calibration of the momentum and step size-related hyperparameters;  developing an adaptive momentum for MomentumRNN is of great research interest.
Finally, we claim that this paper does not have any ethical issue or leverage biases in data.

\section{Acknowledgement}
This material is based on research sponsored by the NSF grant DMS-1924935 and DMS-1952339, and the DOE grant  DE-SC0021142. Other grants that support the work include the NSF grants CCF-1911094, IIS-1838177, and IIS-1730574; the ONR grants N00014-18-12571 and N00014-17-1-2551; the AFOSR grant FA9550-18-1-0478; the DARPA grant G001534-7500; and a Vannevar Bush Faculty Fellowship, ONR grant N00014-18-1-2047.

This material is also based upon work supported by the NSF under Grant\# 2030859 to the Computing Research Association for the CIFellows Project, the NSF Graduate Research Fellowship Program, and the NSF IGERT Training Grant (DGE-1250104). 





\begin{thebibliography}{10}

\bibitem{arjovsky2016unitary}
Martin Arjovsky, Amar Shah, and Yoshua Bengio.
\newblock Unitary evolution recurrent neural networks.
\newblock In {\em International Conference on Machine Learning}, pages
  1120--1128, 2016.

\bibitem{beck2009fast}
Amir Beck and Marc Teboulle.
\newblock A fast iterative shrinkage-thresholding algorithm for linear inverse
  problems.
\newblock {\em SIAM Journal on Imaging Sciences}, 2(1):183--202, 2009.

\bibitem{bengio2013advances}
Yoshua Bengio, Nicolas Boulanger-Lewandowski, and Razvan Pascanu.
\newblock Advances in optimizing recurrent networks.
\newblock In {\em 2013 IEEE International Conference on Acoustics, Speech and
  Signal Processing}, pages 8624--8628. IEEE, 2013.

\bibitem{bengio1994learning}
Yoshua Bengio, Patrice Simard, and Paolo Frasconi.
\newblock Learning long-term dependencies with gradient descent is difficult.
\newblock {\em IEEE Transactions on Neural Networks}, 5(2):157--166, 1994.

\bibitem{dtrivgithub}
Mario~Lezcano Casado.
\newblock Optimization with orthogonal constraints and on general manifolds.
\newblock \url{https://github.com/Lezcano/expRNN}, 2019.

\bibitem{casado2019trivializations}
Mario~Lezcano Casado.
\newblock Trivializations for gradient-based optimization on manifolds.
\newblock In {\em Advances in Neural Information Processing Systems}, pages
  9154--9164, 2019.

\bibitem{chalasani2013fast}
Rakesh Chalasani, Jose~C Principe, and Naveen Ramakrishnan.
\newblock A fast proximal method for convolutional sparse coding.
\newblock In {\em The 2013 International Joint Conference on Neural Networks
  (IJCNN)}, pages 1--5. IEEE, 2013.

\bibitem{chandar2019towards}
Sarath Chandar, Chinnadhurai Sankar, Eugene Vorontsov, Samira~Ebrahimi Kahou,
  and Yoshua Bengio.
\newblock Towards non-saturating recurrent units for modelling long-term
  dependencies.
\newblock In {\em Proceedings of the AAAI Conference on Artificial
  Intelligence}, volume~33, pages 3280--3287, 2019.

\bibitem{chang2019antisymmetricrnn}
Bo~Chang, Minmin Chen, Eldad Haber, and Ed~H Chi.
\newblock Antisymmetricrnn: A dynamical system view on recurrent neural
  networks.
\newblock {\em arXiv preprint arXiv:1902.09689}, 2019.

\bibitem{chen2019symplectic}
Zhengdao Chen, Jianyu Zhang, Martin Arjovsky, and L{\'e}on Bottou.
\newblock Symplectic recurrent neural networks.
\newblock {\em arXiv preprint arXiv:1909.13334}, 2019.

\bibitem{cho2014learning}
Kyunghyun Cho, Bart Van~Merri{\"e}nboer, Caglar Gulcehre, Dzmitry Bahdanau,
  Fethi Bougares, Holger Schwenk, and Yoshua Bengio.
\newblock Learning phrase representations using rnn encoder-decoder for
  statistical machine translation.
\newblock {\em arXiv preprint arXiv:1406.1078}, 2014.

\bibitem{coffey2012langevin}
William Coffey and Yu~P Kalmykov.
\newblock {\em The Langevin equation: with applications to stochastic problems
  in physics, chemistry and electrical engineering}, volume~27.
\newblock World Scientific, 2012.

\bibitem{duane1987hybrid}
Simon Duane, Anthony~D Kennedy, Brian~J Pendleton, and Duncan Roweth.
\newblock Hybrid monte carlo.
\newblock {\em Physics Letters B}, 195(2):216--222, 1987.

\bibitem{elman1990finding}
Jeffrey~L Elman.
\newblock Finding structure in time.
\newblock {\em Cognitive Science}, 14(2):179--211, 1990.

\bibitem{fernandez2007sequence}
Santiago Fern{\'a}ndez, Alex Graves, and J{\"u}rgen Schmidhuber.
\newblock Sequence labelling in structured domains with hierarchical recurrent
  neural networks.
\newblock In {\em Proceedings of the 20th International Joint Conference on
  Artificial Intelligence, IJCAI 2007}, 2007.

\bibitem{garofolo1993timit}
John~S Garofolo.
\newblock Timit acoustic phonetic continuous speech corpus.
\newblock {\em Linguistic Data Consortium, 1993}, 1993.

\bibitem{gers2001lstm}
Felix~A Gers and E~Schmidhuber.
\newblock {LSTM recurrent networks learn simple context-free and
  context-sensitive languages}.
\newblock {\em IEEE Transactions on Neural Networks}, 12(6):1333--1340, 2001.

\bibitem{gers2000recurrent}
Felix~A Gers and J{\"u}rgen Schmidhuber.
\newblock Recurrent nets that time and count.
\newblock In {\em Proceedings of the IEEE-INNS-ENNS International Joint
  Conference on Neural Networks. IJCNN 2000. Neural Computing: New Challenges
  and Perspectives for the New Millennium}, volume~3, pages 189--194. IEEE,
  2000.

\bibitem{gers1999learning}
Felix~A Gers, J{\"u}rgen Schmidhuber, and Fred Cummins.
\newblock Learning to forget: Continual prediction with lstm.
\newblock 1999.

\bibitem{goh2017momentum}
Gabriel Goh.
\newblock Why momentum really works.
\newblock {\em Distill}, 2(4):e6, 2017.

\bibitem{he2019momentum}
Kaiming He, Haoqi Fan, Yuxin Wu, Saining Xie, and Ross Girshick.
\newblock Momentum contrast for unsupervised visual representation learning.
\newblock {\em arXiv preprint arXiv:1911.05722}, 2019.

\bibitem{pmlr-v80-helfrich18a}
Kyle Helfrich, Devin Willmott, and Qiang Ye.
\newblock Orthogonal recurrent neural networks with scaled {C}ayley transform.
\newblock In Jennifer Dy and Andreas Krause, editors, {\em Proceedings of the
  35th International Conference on Machine Learning}, volume~80 of {\em
  Proceedings of Machine Learning Research}, pages 1969--1978,
  Stockholmsmässan, Stockholm Sweden, 10--15 Jul 2018. PMLR.

\bibitem{pmlr-v48-henaff16}
Mikael Henaff, Arthur Szlam, and Yann LeCun.
\newblock Recurrent orthogonal networks and long-memory tasks.
\newblock In Maria~Florina Balcan and Kilian~Q. Weinberger, editors, {\em
  Proceedings of The 33rd International Conference on Machine Learning},
  volume~48 of {\em Proceedings of Machine Learning Research}, pages
  2034--2042, New York, New York, USA, 20--22 Jun 2016. PMLR.

\bibitem{hochreiter1997long}
Sepp Hochreiter and J{\"u}rgen Schmidhuber.
\newblock Long short-term memory.
\newblock {\em Neural Computation}, 9(8):1735--1780, 1997.

\bibitem{jing2017tunable}
Li~Jing, Yichen Shen, Tena Dubcek, John Peurifoy, Scott Skirlo, Yann LeCun, Max
  Tegmark, and Marin Solja{\v{c}}i{\'c}.
\newblock Tunable efficient unitary neural networks (eunn) and their
  application to rnns.
\newblock In {\em Proceedings of the 34th International Conference on Machine
  Learning-Volume 70}, pages 1733--1741. JMLR. org, 2017.

\bibitem{kag2019rnns}
Anil Kag, Ziming Zhang, and Venkatesh Saligrama.
\newblock {RNNs} evolving in equilibrium: A solution to the vanishing and
  exploding gradients.
\newblock {\em arXiv preprint arXiv:1908.08574}, 2019.

\bibitem{kamilov2016learning}
US~Kamilov and H~Mansour.
\newblock Learning mmse optimal thresholds for fista.
\newblock 2016.

\bibitem{kingma2014adam}
Diederik~P Kingma and Jimmy Ba.
\newblock Adam: A method for stochastic optimization.
\newblock {\em arXiv preprint arXiv:1412.6980}, 2014.

\bibitem{kotsias2019direct}
Panagiotis-Christos Kotsias, Josep Ar{\'u}s-Pous, Hongming Chen, Ola Engkvist,
  Christian Tyrchan, and Esben~Jannik Bjerrum.
\newblock Direct steering of de novo molecular generation using descriptor
  conditional recurrent neural networks (crnns).
\newblock 2019.

\bibitem{kusupati2018fastgrnn}
Aditya Kusupati, Manish Singh, Kush Bhatia, Ashish Kumar, Prateek Jain, and
  Manik Varma.
\newblock Fastgrnn: A fast, accurate, stable and tiny kilobyte sized gated
  recurrent neural network.
\newblock In {\em Advances in Neural Information Processing Systems}, pages
  9017--9028, 2018.

\bibitem{laurent2016recurrent}
Thomas Laurent and James von Brecht.
\newblock A recurrent neural network without chaos.
\newblock {\em arXiv preprint arXiv:1612.06212}, 2016.

\bibitem{le2015simple}
Quoc~V Le, Navdeep Jaitly, and Geoffrey~E Hinton.
\newblock A simple way to initialize recurrent networks of rectified linear
  units.
\newblock {\em arXiv preprint arXiv:1504.00941}, 2015.

\bibitem{lecun2010mnist}
Yann LeCun, Corinna Cortes, and CJ~Burges.
\newblock {MNIST handwritten digit database}.
\newblock {\em ATT Labs [Online]. Available: http://yann.lecun.com/exdb/mnist},
  2, 2010.

\bibitem{lezcano2019cheap}
Mario Lezcano-Casado and David Mart{\'i}nez-Rubio.
\newblock Cheap orthogonal constraints in neural networks: A simple
  parametrization of the orthogonal and unitary group.
\newblock In {\em International Conference on Machine Learning (ICML)}, pages
  3794--3803, 2019.

\bibitem{li2018independently}
Shuai Li, Wanqing Li, Chris Cook, Ce~Zhu, and Yanbo Gao.
\newblock Independently recurrent neural network (indrnn): Building a longer
  and deeper rnn.
\newblock In {\em Proceedings of the IEEE conference on computer vision and
  pattern recognition}, pages 5457--5466, 2018.

\bibitem{mccann2017convolutional}
Michael~T McCann, Kyong~Hwan Jin, and Michael Unser.
\newblock Convolutional neural networks for inverse problems in imaging: A
  review.
\newblock {\em IEEE Signal Processing Magazine}, 34(6):85--95, 2017.

\bibitem{merity2018regularizing}
Stephen Merity, Nitish~Shirish Keskar, and Richard Socher.
\newblock Regularizing and optimizing {LSTM} language models.
\newblock In {\em International Conference on Learning Representations}, 2018.

\bibitem{mhammedi2017efficient}
Zakaria Mhammedi, Andrew Hellicar, Ashfaqur Rahman, and James Bailey.
\newblock Efficient orthogonal parametrisation of recurrent neural networks
  using householder reflections.
\newblock In {\em Proceedings of the 34th International Conference on Machine
  Learning-Volume 70}, pages 2401--2409. JMLR. org, 2017.

\bibitem{mikolov2010recurrent}
Tom{\'a}{\v{s}} Mikolov, Martin Karafi{\'a}t, Luk{\'a}{\v{s}} Burget, Jan
  {\v{C}}ernock{\`y}, and Sanjeev Khudanpur.
\newblock Recurrent neural network based language model.
\newblock In {\em Eleventh Annual Conference of the International Speech
  Communication Association}, 2010.

\bibitem{moreau2017understanding}
Thomas Moreau and Joan Bruna.
\newblock Understanding the learned iterative soft thresholding algorithm with
  matrix factorization.
\newblock {\em arXiv preprint arXiv:1706.01338}, 2017.

\bibitem{neal2011mcmc}
Radford~M Neal et~al.
\newblock {MCMC} using {Hamiltonian} dynamics.

\bibitem{neil2016phased}
Daniel Neil, Michael Pfeiffer, and Shih-Chii Liu.
\newblock {Phased LSTM: Accelerating recurrent network training for long or
  event-based sequences}.
\newblock In {\em Advances in Neural Information Processing Systems}, pages
  3882--3890, 2016.

\bibitem{nemirovskii1985optimal}
Arkaddii~S Nemirovskii and Yu~E Nesterov.
\newblock Optimal methods of smooth convex minimization.
\newblock {\em USSR Computational Mathematics and Mathematical Physics},
  25(2):21--30, 1985.

\bibitem{nesterov1983method}
Yurii~E Nesterov.
\newblock A method for solving the convex programming problem with convergence
  rate o (1/k\^{} 2).
\newblock In {\em Dokl. Akad. Nauk Sssr}, volume 269, pages 543--547, 1983.

\bibitem{niu2019recurrent}
Murphy~Yuezhen Niu, Lior Horesh, and Isaac Chuang.
\newblock Recurrent neural networks in the eye of differential equations.
\newblock {\em arXiv preprint arXiv:1904.12933}, 2019.

\bibitem{palangi2016deep}
Hamid Palangi, Li~Deng, Yelong Shen, Jianfeng Gao, Xiaodong He, Jianshu Chen,
  Xinying Song, and Rabab Ward.
\newblock Deep sentence embedding using long short-term memory networks:
  Analysis and application to information retrieval.
\newblock {\em IEEE/ACM Transactions on Audio, Speech, and Language
  Processing}, 24(4):694--707, 2016.

\bibitem{pascanu2013difficulty}
Razvan Pascanu, Tomas Mikolov, and Yoshua Bengio.
\newblock On the difficulty of training recurrent neural networks.
\newblock In {\em International Conference on Machine Learning}, pages
  1310--1318, 2013.

\bibitem{paszke2019pytorch}
Adam Paszke, Sam Gross, Francisco Massa, Adam Lerer, James Bradbury, Gregory
  Chanan, Trevor Killeen, Zeming Lin, Natalia Gimelshein, Luca Antiga, et~al.
\newblock Pytorch: An imperative style, high-performance deep learning library.
\newblock In {\em Advances in Neural Information Processing Systems}, pages
  8024--8035, 2019.

\bibitem{polyak1964some}
Boris~T Polyak.
\newblock Some methods of speeding up the convergence of iteration methods.
\newblock {\em USSR Computational Mathematics and Mathematical Physics},
  4(5):1--17, 1964.

\bibitem{pulver2017lstm}
Andrew Pulver and Siwei Lyu.
\newblock {LSTM with working memory}.
\newblock In {\em 2017 International Joint Conference on Neural Networks
  (IJCNN)}, pages 845--851. IEEE, 2017.

\bibitem{qu2017syllable}
Zhongdi Qu, Parisa Haghani, Eugene Weinstein, and Pedro Moreno.
\newblock {Syllable-based acoustic modeling with CTC-SMBR-LSTM}.
\newblock In {\em 2017 IEEE Automatic Speech Recognition and Understanding
  Workshop (ASRU)}, pages 173--177. IEEE, 2017.

\bibitem{rahman2016new}
Lamia Rahman, Nabeel Mohammed, and Abul~Kalam Al~Azad.
\newblock {A new LSTM model by introducing biological cell state}.
\newblock In {\em 2016 3rd International Conference on Electrical Engineering
  and Information Communication Technology (ICEEICT)}, pages 1--6. IEEE, 2016.

\bibitem{sak2014long}
Ha{\c{s}}im Sak, Andrew Senior, and Fran{\c{c}}oise Beaufays.
\newblock Long short-term memory based recurrent neural network architectures
  for large vocabulary speech recognition.
\newblock {\em arXiv preprint arXiv:1402.1128}, 2014.

\bibitem{ptbgithub}
Salesforce.
\newblock Lstm and qrnn language model toolkit for pytorch.
\newblock \url{https://github.com/salesforce/awd-lstm-lm}, 2017.

\bibitem{sutskever2013importance}
Ilya Sutskever, James Martens, George Dahl, and Geoffrey Hinton.
\newblock On the importance of initialization and momentum in deep learning.
\newblock In {\em International Conference on Machine Learning}, pages
  1139--1147, 2013.

\bibitem{szlam2011structured}
Arthur~D Szlam, Karol Gregor, and Yann~L Cun.
\newblock Structured sparse coding via lateral inhibition.
\newblock In {\em Advances in Neural Information Processing Systems}, pages
  1116--1124, 2011.

\bibitem{talathi2015improving}
Sachin~S Talathi and Aniket Vartak.
\newblock Improving performance of recurrent neural network with relu
  nonlinearity.
\newblock {\em arXiv preprint arXiv:1511.03771}, 2015.

\bibitem{Tieleman2012}
T.~Tieleman and G.~Hinton.
\newblock {Lecture 6.5---RmsProp: Divide the gradient by a running average of
  its recent magnitude}.
\newblock COURSERA: Neural Networks for Machine Learning, 2012.

\bibitem{van2018unreasonable}
Jos Van Der~Westhuizen and Joan Lasenby.
\newblock The unreasonable effectiveness of the forget gate.
\newblock {\em arXiv preprint arXiv:1804.04849}, 2018.

\bibitem{vorontsov2017orthogonality}
Eugene Vorontsov, Chiheb Trabelsi, Samuel Kadoury, and Chris Pal.
\newblock On orthogonality and learning recurrent networks with long term
  dependencies.
\newblock In {\em Proceedings of the 34th International Conference on Machine
  Learning-Volume 70}, pages 3570--3578. JMLR. org, 2017.

\bibitem{wang2020scheduled}
Bao Wang, Tan~M Nguyen, Andrea~L Bertozzi, Richard~G Baraniuk, and Stanley~J
  Osher.
\newblock Scheduled restart momentum for accelerated stochastic gradient
  descent.
\newblock {\em arXiv preprint arXiv:2002.10583}, 2020.

\bibitem{wisdom2016full}
Scott Wisdom, Thomas Powers, John Hershey, Jonathan Le~Roux, and Les Atlas.
\newblock Full-capacity unitary recurrent neural networks.
\newblock In {\em Advances in Neural Information Processing Systems}, pages
  4880--4888, 2016.

\end{thebibliography}

\clearpage

\appendix
\begin{center}
{\bf Appendix for "MomentumRNN: Integrating Momentum into Recurrent Neural Networks"}
\end{center}



\section{Experimental Details}\label{sec:exp:details}
In this section, we describe the datasets used in our experiments and provide details on the model implementation and training. MomentumLSTM, AdamLSTM, RMSPropLSTM, and SRLSTM, as well as MomentumDTRIV, AdamDTRIV, RMSPropDTRIV, and SRDTRIV share the same settings as their LSTM/DTRIV counterparts with the additional momentum $\mu$, step size $s$,  scheduled restart $F$, and the coefficient $\beta$ used for computing running averages of the squared gradients. Thus, we only provide implementation and training details for the baseline LSTM and DTRIV for each task. Values for additional hyperparameters in our momentum-based models are found by grid search and reported in Table~\ref{tab:mlstm-hyperparams},~\ref{tab:alstm-hyperparams},~\ref{tab:rlstm-hyperparams}, and~\ref{tab:slstm-hyperparams}.


\subsection{Pixel-by-Pixel MNIST}
\label{sec:mnist-appendix}
MNIST dataset~\cite{lecun2010mnist} consists of 60K training images and 10K test images from 10 classes of hand-written digits. Both training and test data are binary images of size $28 \times 28$. As mentioned in Section~\ref{sec:mnist-exp}, we flatten and process the image as a sequence of the length of 784 pixel-by-pixel. In the unpermuted task (MNIST), the images are processed row-by-row, while in the permuted task (PMNIST), a fixed permutation is applied to both training and test images.

{\bf LSTM.} The baseline LSTM models consist of one LSTM cell with 128 and 256 hidden units. Orthogonal initialization is used for input-to-hidden weights, while hidden-to-hidden weights are initialized to identity matrices. The forget gate bias is initialized to 1 while all other bias scalars are initialized to 0. We follow LSTM training in~\cite{lezcano2019cheap, casado2019trivializations} to train LSTM models for the MNIST and PMNIST tasks. Gradient norms are clipped to 1 during training, and the smoothing constant $\alpha$ for the RMSProp optimizer is set to $0.9$. We provide other details on hyperparameters for the LSTM training on (P)MNIST in Table~\ref{tab:lstm-hyperparams} (top).

{\bf DTRIV.} We use the best DTRIV models for each (P)MNIST task reported in~\cite{casado2019trivializations} with Cayley initialization~\cite{pmlr-v80-helfrich18a}. The gradient norms are clipped to 1 during training. Other hyperparameter details are provided in Table~\ref{tab:lstm-hyperparams} (bottom). 

\subsection{TIMIT Speech Dataset}
TIMIT speech dataset is a collection of real-world speech recordings~\cite{garofolo1993timit} consisting of 3640 utterances for the training set, 192 utterances for the validation set, and 400 utterances for the test set. We follow the data preprocessing in~\cite{wisdom2016full, casado2019trivializations, lezcano2019cheap, pmlr-v80-helfrich18a}. In particular, audio files in TIMIT are downsampled to 8kHz. A short-time Fourier transform (STFT) is then applied with a Hann window of 256 samples and a window hop of 128 samples (16 milliseconds) to yield sequences of 129 complex-valued Fourier amplitudes. The log-magnitude of these sequences is fed into the models as the input data. The task is to predict the next log-magnitude given the previous ones.    

{\bf LSTM.} The baseline LSTM models consist of one LSTM cell with 84, 120, and 158 hidden units. Similar to (P)MNIST experiments, orthogonal initialization is used for input-to-hidden weights, while hidden-to-hidden weights are initialized to identity matrices. However, the forget gate bias is initialized to -4 while all other bias scalars are initialized to 0. We follow LSTM training in~\cite{lezcano2019cheap, casado2019trivializations} to train LSTM models for the TIMIT tasks. We use the standard Adam optimizer in PyTorch~\cite{paszke2019pytorch} to train the models without using gradient clipping. We provide other details on hyperparameters for the LSTM training on TIMIT in Table~\ref{tab:lstm-hyperparams} (top).

{\bf DTRIV.} We use the best DTRIV models for each TIMIT task reported in~\cite{casado2019trivializations} with Henaff initialization~\cite{pmlr-v48-henaff16}. Other hyperparameter details are provided in Table~\ref{tab:lstm-hyperparams} (bottom). 

\subsection{Word-Level Penn TreeBank}
The Penn TreeBank (PTB) dataset is among the most popular datasets for experimenting with language modeling. The dataset has 10,000 unique words and is preprocessed to not include capital letters, numbers, or punctuation~\cite{mikolov2010recurrent}. 

{\bf LSTM.} The baseline are three-layer LSTM models with 1150 hidden units at each layer and an embedding of size 400. We follow the LSTM implementation and training in~\cite{merity2018regularizing}. We summarize some important details in Table~\ref{tab:lstm-hyperparams} (top).

\subsection{Copying and Adding Tasks}
\label{sec:copying-adding-appendix}

We define the copying and adding tasks in Section~\ref{sec:copying-adding-maintext} as follows.

{\bf Copying task. } In the copying task, we consider a set $A$ of $N$ alphabet, e.g. $A = \{a_k\}_{k=1}^{N}$, and let $\code{<start>}$ and $\code{<blank>}$ be two symbols not contained in $A$. 
For a sequence 
\begin{wrapfigure}[9]{r}{.4\textwidth}
\vspace{-0.17 in}
\includegraphics[width=\linewidth]{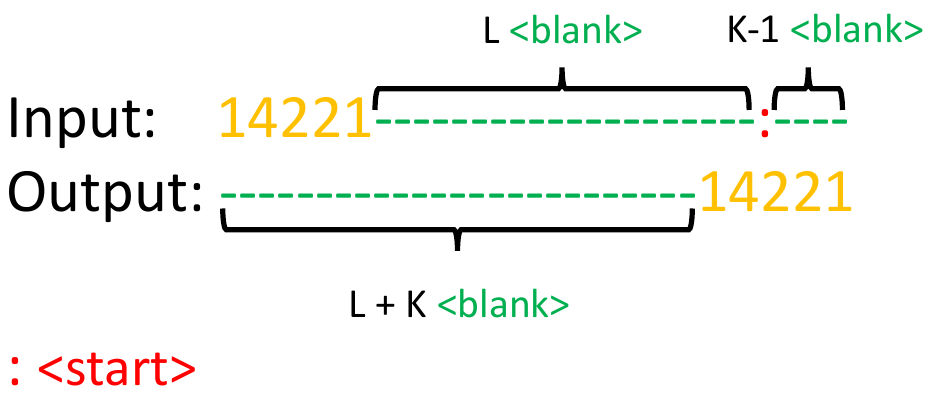}
\vspace{-0.2 in}
\caption{An example of input and output in the copying task.}
\label{fig:copy_task_example}
\end{wrapfigure}
of $K$ ordered characters sampled i.i.d. uniformly from A 
and a spacing length L, the input sequence is the $K$ characters followed by $L$ $\code{<blank>}$ characters, a $\code{<start>}$ character, and then $K - 1$ $\code{<blank>}$ characters. The task is to output a sequence containing $K + L$ $\code{<blank>}$ characters followed by the alphabet character sequence of length $K$. For example, let $A = \{1,2,3,4\}$, $K=5$, $L=20$, $\code{<start>}=:$, and $\code{<blank>}=-$, an input sequence and its corresponding output sequence is given in Figure~\ref{fig:copy_task_example}.

{\bf Adding task. } We follow the adding problem as proposed
in~\cite{arjovsky2016unitary}, which is a variation of the similar problem in~\cite{hochreiter1997long}. In particular, in this task, two sequences of length T are concurrently passed into an RNN. The first sequence consists of ordered digits sampled uniformly from a half-open interval $U[0, 1]$. The second sequence contains all zeros except for two entries that are marked by 1. The location of the first and second 1 is uniformly chosen within the interval $[1, T/2]$ and $[T/2, T]$, respectively. We label each pair of sequences by the sum of the two entries in the first sequence that are marked by 1's in the second sequence. 

{\bf LSTM.} The baseline LSTM models for the copying and adding tasks consist of one LSTM cell with 190 and 128 hidden units, respectively. Orthogonal initialization is used for input-to-hidden weights, while hidden-to-hidden weights are initialized to identity matrices. The forget gate bias is initialized to 1 while all other bias scalars are initialized to 0. We follow LSTM training in~\cite{lezcano2019cheap} and~\cite{li2018independently} to train LSTM models for the copying and adding tasks, respectively. We provide details on hyperparameters for the LSTM training on the copying task in Table~\ref{tab:lstm-hyperparams} (top).

\subsection{Momentum Cells can Avoid Vanishing Gradient Issue}\label{sec:appendix:stability}
To confirm that MomentumRNN can alleviate vanishing gradients, we train a MomentumDTRIV and its corresponding baseline DTRIV for the PMNIST classification task. We plot $\|\partial \mathcal L/\partial \vh_t\|_{2}$ for each time step $t$ at each training iteration, as shown in Figure~\ref{fig:stability}. Both MomentumDTRIV and DTRIV models used in this experiment contains one cell of 170 hidden units. The model implementation and training details are similar to those in Section~\ref{sec:mnist-appendix} above. Note that DTRIV is also an RNN with additional orthogonality constraint. 

\section{Backpropagation Through Time -- A Review}\label{sec:appendix:BPTT}
In this section, we give a short review of the backpropagation through time, which is a major algorithm for training RNNs. We consider the standard recurrent cell~\eqref{eq:RNN:Cell}, and for any given training sample $(\vx, \vy)$ with $\vx = (\vx_1, \cdots, \vx_T)$ being an input sequence of length $T$ and $\vy = (y_1, \cdots, y_T)$ being the sequence of labels \footnote{Without loss of generality, we consider the sequence to sequence modeling.}. Let $\mathcal{L}_t$ be the loss at the time step $t$ and the total loss on the whole sequence is
\begin{equation}
\label{eq:whole:loss}
\mathcal{L} = \sum_{t=1}^T \mathcal{L}_t.
\end{equation}
For any $1\leq t\leq T$, we can compute the gradient of the loss $\mathcal{L}_t$ with respect to the parameter $\Ub$ as
\begin{equation}
\label{eq:gradient:U}
\frac{\partial \mathcal{L}_t}{\partial \Ub} = \sum_{k=1}^t \frac{\partial \vh_k}{\partial \Ub}\cdot \frac{\partial \mathcal{L}_t}{\partial \vh_t}\cdot \frac{\partial \vh_t}{\partial \vh_k} = \sum_{k=1}^t \frac{\partial\vh_k}{\partial \Ub} \cdot \frac{\partial \mathcal{L}_t}{\partial \vh_t} \cdot \prod_{k=1}^{t-1}\frac{\partial\vh_{k+1}}{\partial\vh_k},
\end{equation}
where $\frac{\partial\vh_{k+1}}{\partial\vh_k} = \Db_k\Ub^{\rm T}$ with $\Db_k = {\rm diag}(\sigma'(\Ub\vh_k + \Wb\vx_{k+1} + \vb))$. Similarly, we can compute $\partial \mathcal{L}_t/\partial \Wb$ and $\partial \mathcal{L}_t/\partial \vb$.

\begin{table}[!h]
\vspace{-0.1in}
    \caption{Hyperparameters for the Baseline LSTM and DTRIV Training.}
\vspace{-0.1in}
\label{tab:lstm-hyperparams}
\begin{center}
{\centering LSTM\\}
\vspace{0.1 in}
{\tiny
\begin{tabular}{l|cccc}
    \toprule
    Dataset & Optimizer & Learning Rate & Batch Size & \#Epochs \\
    \midrule
    \midrule
    MNIST & RMSProp & $0.001$ & $128$ & $150$ \\
    PMNIST & RMSProp & $0.001$ & $128$ & $150$ \\
    TIMIT & Adam & $0.0001$ & $32$ & $700$ \\
    PTB & SGD & $30$ (initial learning rate) & $20$ & $500$ \\
    Copying & RMSprop & $0.0002$ & $128$ & $7000$ \\
    Adding & Adam & $0.0002$ & $50$ & $1200$ \\
    \bottomrule
\end{tabular}
}
\\
\vspace{0.1 in}
{\centering DTRIV\\}
\vspace{0.1 in}
{\tiny
\begin{tabular}{lc|ccccccc}
    \toprule
    Dataset & Size & DTRIV & Optimizer & Learning & Orthogonal & Orthogonal & Batch & \#Epochs \\
     &  & Opt. & & Rate & Optimizer & Learning & Size & \\
     &  & Step (K) & & & & Rate & & \\
    \midrule
    \midrule
    MNIST & 170 & $1$ & & $0.001$ & & $0.0001 $ & $128$ & $150$ \\
    MNIST & 360 & $\infty$ & RMSProp & $0.0005$ & RMSProp & $0.0001$ & $128$ & $150$ \\
    MNIST & 512 & $100$ & & $0.0005$ & & $0.0001$ &  $128$ & $150$ \\
    \midrule
    PMNIST & 170 & $1$ & & $0.0007$ & & $0.0002 $ & $128$ & $150$ \\
    PMNIST & 360 & $\infty$ & RMSProp & $0.0007$ & RMSProp & $0.00005$ & $128$ & $150$ \\
    PMNIST & 512 & $\infty$ & & $0.0003$ & & $0.00007$ &  $128$ & $150$ \\
    \midrule
    TIMIT & 224 & $\infty$ & Adam & $0.001$ & RMSProp & $0.0002$ & $128$ & $700$ \\
    TIMIT & 322 & $\infty$ & & $0.001$ & & $0.0002$ & $128$ & $700$ \\
    \bottomrule
\end{tabular}
}
\end{center}
\end{table}

\begin{table}[!h]
    \caption{Hyperparameters for MomentumLSTM and MomentumDTRIV Training}
\vspace{-0.1in}
\label{tab:mlstm-hyperparams}
\begin{center}
{\centering MomentumLSTM\\}
\vspace{0.1 in}
{\tiny
\begin{tabular}{l|cccccc}
    \toprule
    Dataset & Momentum $\mu$ & Step Size $s$ & Optimizer & Learning Rate & Batch Size & \#Epochs \\
    \midrule
    \midrule
    MNIST & $0.6$ & $0.6$ & RMSProp & $0.001$ & $128$ & $150$ \\
    PMNIST & $0.6$ & $1.0$ & RMSProp & $0.001$ & $128$ & $150$ \\
    TIMIT & $0.3$ & $0.1$ & Adam & $0.0001$ & $32$ & $700$ \\
    PTB & $0.0$ & $0.6$ & SGD & $30$ (initial learning rate) & $20$ & $500$ \\
    Copying (sequence length 1K) & $0.6$ & $0.9$ & RMSprop & $0.0002$ & $128$ & $7000$ \\
    Copying (sequence length 2K) & $0.9$ & $2.0$ & RMSprop & $0.0002$ & $128$ & $7000$ \\
    Adding & $0.9$ & $2.0$ & Adam & $0.0002$ & $50$ & $1200$ \\
    \bottomrule
\end{tabular}
}
\\
\vspace{0.1 in}
{\centering MomentumDTRIV\\}
\vspace{0.1 in}
{\tiny
\begin{tabular}{lc|ccccccccc}
    \toprule
    Dataset & Size & DTRIV & Momentum & Step Size & Optimizer & Learning & Orthogonal & Orthogonal & Batch & \#Epochs \\
     &  & Opt. & $\mu$ & $s$ & & Rate & Optimizer & Learning & Size & \\
     &  & Step (K) & & & & & & Rate & & \\
    \midrule
    \midrule
    PMNIST & 170 & $1$ & $0.6$ & $0.9$ & & $0.0007$ & & $0.0002 $ & $128$ & $150$ \\
    PMNIST & 360 & $\infty$ & $0.3$ & $0.3$ & RMSProp & $0.0007$ & RMSProp & $0.00005$ & $128$ & $150$ \\
    PMNIST & 512 & $\infty$ & $0.3$ & $0.3$ & & $0.0003$ & & $0.00007$ &  $128$ & $150$ \\
    \midrule
    TIMIT & 224 & $\infty$ & $0.3$ & $0.1$ & Adam & $0.001$ & RMSProp & $0.0002$ & $128$ & $700$ \\
    TIMIT & 322 & $\infty$ & $0.3$ & $0.1$ & & $0.001$ & & $0.0002$ & $128$ & $700$ \\
    \bottomrule
\end{tabular}
}
\end{center}
\end{table}

\begin{table}[!h]
    \caption{Hyperparameters for AdamLSTM and AdamDTRIV Training}
\vspace{-0.1in}
\label{tab:alstm-hyperparams}
\begin{center}
{\centering AdamLSTM\\}
\vspace{0.1 in}
{\tiny
\begin{tabular}{l|ccccccc}
    \toprule
    Dataset & Optimizer & Momentum $\mu$ & Step Size $s$ & $\beta$ & Learning Rate & Batch Size & \#Epochs \\
    \midrule
    \midrule
    MNIST & RMSProp & $0.6$ & $0.6$ & $0.1$ & $0.001$ & $128$ & $150$ \\
    PMNIST & RMSProp & $0.6$ & $1.0$ & $0.01$ & $0.001$ & $128$ & $150$ \\
    TIMIT & Adam & $0.3$ & $0.1$ & $0.999$ & $0.0001$ & $32$ & $700$ \\
    Copying (sequence length 1K) & RMSprop & $0.6$ & $2.0$ & $0.999$ & $0.0002$ & $128$ & $7000$ \\
    Copying (sequence length 2K) & RMSprop & $0.6$ & $2.0$ & $0.999$ & $0.0002$ & $128$ & $7000$ \\
    Adding & Adam & $0.6$ & $2.0$ & $0.999$ & $0.0002$ & $50$ & $1200$ \\
    \bottomrule
\end{tabular}
}
\\
\vspace{0.1 in}
{\centering AdamDTRIV\\}
\vspace{0.1 in}
{\tiny
\begin{tabular}{lc|cccccccccc}
    \toprule
    Dataset & Size & DTRIV & Momentum & Step Size & $\beta$ & Optimizer & Learning & Orthogonal & Orthogonal & Batch & \#Epochs \\
     &  & Opt. & $\mu$ & $s$ & & & Rate & Optimizer & Learning & Size & \\
     &  & Step (K) & & & & & & & Rate & & \\
    \midrule
    \midrule
    PMNIST & 512 & $\infty$ & $0.3$ & $0.3$ & $0.8$ & RMSProp & $0.0003$ & RMSProp & $0.00007$ &  $128$ & $150$ \\
    \bottomrule
\end{tabular}
}
\end{center}
\vspace{0.5cm}
\end{table}

\begin{table}[!h]
    \caption{Hyperparameters for RMSPropLSTM and RMSPropDTRIV Training}
\vspace{-0.1in}
\label{tab:rlstm-hyperparams}
\begin{center}
{\centering RMSPropLSTM\\}
\vspace{0.1 in}
{\tiny
\begin{tabular}{l|ccccccc}
    \toprule
    Dataset & Optimizer & Momentum $\mu$ & Step Size $s$ & $\beta$ & Learning Rate & Batch Size & \#Epochs \\
    \midrule
    \midrule
    MNIST & RMSProp & $0.0$ & $0.6$ & $0.9$ (size $N=256$) & $0.001$ & $128$ & $150$ \\
     &  &  &  & $0.99$ (size $N=128$) &  &  &  \\
    PMNIST & RMSProp & $0.0$ & $1.0$ & $0.01$ & $0.001$ & $128$ & $150$ \\
    TIMIT & Adam & $0.0$ & $0.1$ & $0.999$ & $0.0001$ & $32$ & $700$ \\
    Copying (sequence length 1K) & RMSprop & $0.0$ & $2.0$ & $0.999$ & $0.0002$ & $128$ & $7000$ \\
    Copying (sequence length 2K) & RMSprop & $0.0$ & $2.0$ & $0.999$ & $0.0002$ & $128$ & $7000$ \\
    Adding & Adam & $0.0$ & $2.0$ & $0.999$ & $0.0002$ & $50$ & $1200$ \\
    \bottomrule
\end{tabular}
}
\\
\vspace{0.1 in}
{\centering RMSPropDTRIV\\}
\vspace{0.1 in}
{\tiny
\begin{tabular}{lc|cccccccccc}
    \toprule
    Dataset & Size & DTRIV & Momentum & Step Size & $\beta$ & Optimizer & Learning & Orthogonal & Orthogonal & Batch & \#Epochs \\
     &  & Opt. & $\mu$ & $s$ & & & Rate & Optimizer & Learning & Size & \\
     &  & Step (K) & & & & & & & Rate & & \\
    \midrule
    \midrule
    PMNIST & 512 & $\infty$ & $0.0$ & $0.3$ & $0.9$ & RMSProp & $0.0003$ & RMSProp & $0.00007$ &  $128$ & $150$ \\
    \bottomrule
\end{tabular}
}
\end{center}
\vspace{0.5cm}
\end{table}

\begin{table}[!h]
\vspace{-0.8cm}
    \caption{Hyperparameters for SRLSTM and SRDTRIV Training}
\vspace{-0.1in}
\label{tab:slstm-hyperparams}
\begin{center}
{\centering SRLSTM\\}
\vspace{0.1 in}
{\tiny
\begin{tabular}{l|cccccc}
    \toprule
    Dataset & Optimizer & Scheduled & Step Size $s$ & Learning Rate & Batch Size & \#Epochs \\
     & & Restart (F) & & & &  \\
    \midrule
    \midrule
    MNIST & RMSProp & $2$ & $1.0$ & $0.001$ & $128$ & $150$ \\
    PMNIST & RMSProp & $40$ (size $N=256$) & $0.9$ (size $N=256$) & $0.001$ & $128$ & $150$ \\
     & & $6$ (size $N=128$) & $0.01$ (size $N=128$) &  &  &  \\
    TIMIT & Adam & $2$ & $0.1$ & $0.0001$ & $32$ & $700$ \\
    PTB & SGD & $2$ & $0.6$ & $30$ (initial learning rate) & $20$ & $500$ \\
    Copying & RMSprop & $100$ & $0.9$ & $0.0002$ & $128$ & $7000$ \\
    (sequence length 1K) &  &  &  &  &  &  \\
    Copying & RMSprop & $100$ & $0.9$ & $0.0002$ & $128$ & $7000$ \\
    (sequence length 2K) &  &  &  &  &  &  \\
    Adding & Adam & $100$ & $0.9$ & $0.0002$ & $50$ & $1200$ \\
    \bottomrule
\end{tabular}
}
\\
\vspace{0.1 in}
{\centering SRDTRIV\\}
\vspace{0.1 in}
{\tiny
\begin{tabular}{lc|ccccccccc}
    \toprule
    Dataset & Size & DTRIV & Scheduled & Step Size & Optimizer & Learning & Orthogonal & Orthogonal & Batch & \#Epochs \\
     &  & Opt. & Restart (F) & $s$ & & Rate & Optimizer & Learning & Size & \\
     &  & Step (K) & & & & & & Rate & & \\
    \midrule
    \midrule
    PMNIST & 512 & $\infty$ & $2$ & $0.3$ & RMSProp & $0.0003$ & RMSProp & $0.00007$ &  $128$ & $150$ \\
    \bottomrule
\end{tabular}
}
\end{center}
\end{table}

\newpage
\section{More Experimental Results}\label{sec:appendix:more:exp:results}
\begin{figure}[!t]
\centering
\includegraphics[width=1.0\linewidth]{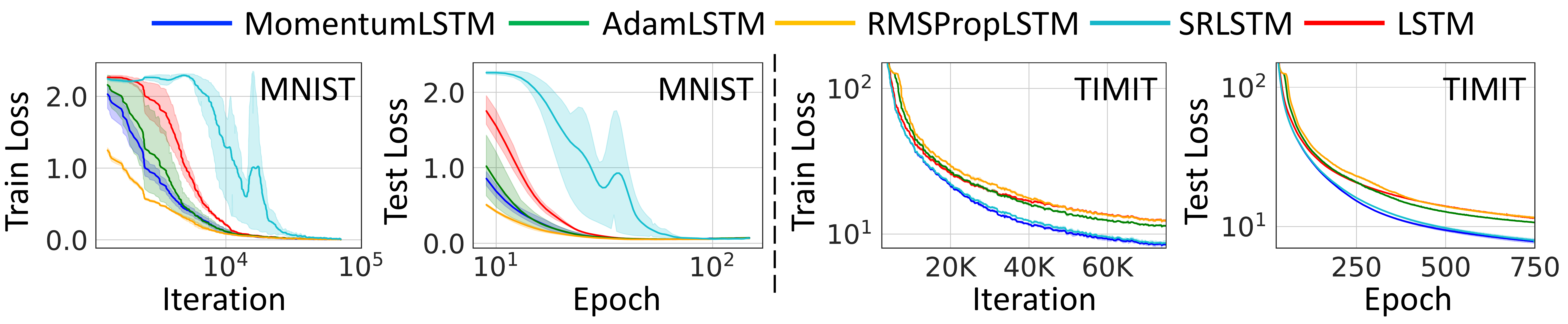}
\vskip -0.2cm
\caption{Train and test loss  of MomentumLSTM (blue), AdamLSTM (green), RMSPropLSTM (orange), SRLSTM (cyan), and LSTM (red) using $N=128$ hidden units for MNIST (left two panels) and using $N=120$ hidden units for TIMIT (right two panels) tasks. MomentumLSTM converges faster than LSTM in both tasks. RMSPropLSTM and MomentumLSTM/SRLSTM converge the fastest for MNIST and TIMIT tasks, respectively.}
\label{fig:loss-vs-iters-appendix}
\end{figure}

We conduct more comprehensive experiments for the Adam principled and NAG principled RNNs. In particular, we perform (P)MNIST and TIMIT experiments using the AdamLSTM, RMSPropLSTM, and SRLSTM of 128 and 120 hidden units, respectively. For (P)MNIST task, RMSPropLSTM achieves the best test accuracy and converges the fastest. For the TIMIT task, MomentumLSTM and SRLSTM outperform the other models while converging faster. We summarize our results in Table~\ref{tab:mnist-appendix-n128} and~\ref{tab:timit-appendix-n120}, as well as in Figure~\ref{fig:loss-vs-iters-appendix}. Note that in the main text, we conduct the same experiments using the same models but with different numbers of hidden units (i.e. 256 hidden units for the (P)MNIST task and 158 hidden units for the TIMIT task).

\begin{figure}[!h]
\centering
\vspace{-0.1 in}
\includegraphics[width=1.0\linewidth]{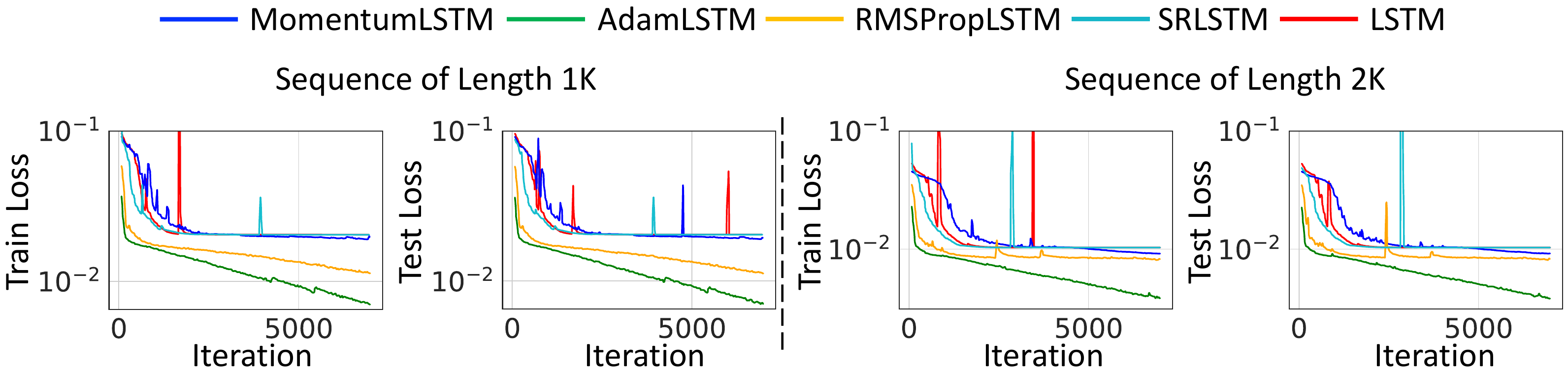}
\vskip -0.2cm
\caption{Train test loss vs. iteration for copying task with sequence length 1K (left) and 2K (right). AdamLSTM and RMSPropLSTM converge faster and to better final losses than other models. MomentumLSTM and SRLSTM converge to similar losses as LSTM.}
\label{fig:adding-appendix}
\vspace{2cm}
\end{figure}

Furthermore, we provide additional results on copying task for sequences of length 1K in comparison with those for sequences of length 2K as in the main text. In addition to training losses, we also include test losses in in Figure~\ref{fig:adding-appendix}.

\begin{figure}[!t]
\centering
\includegraphics[width=0.7\linewidth]{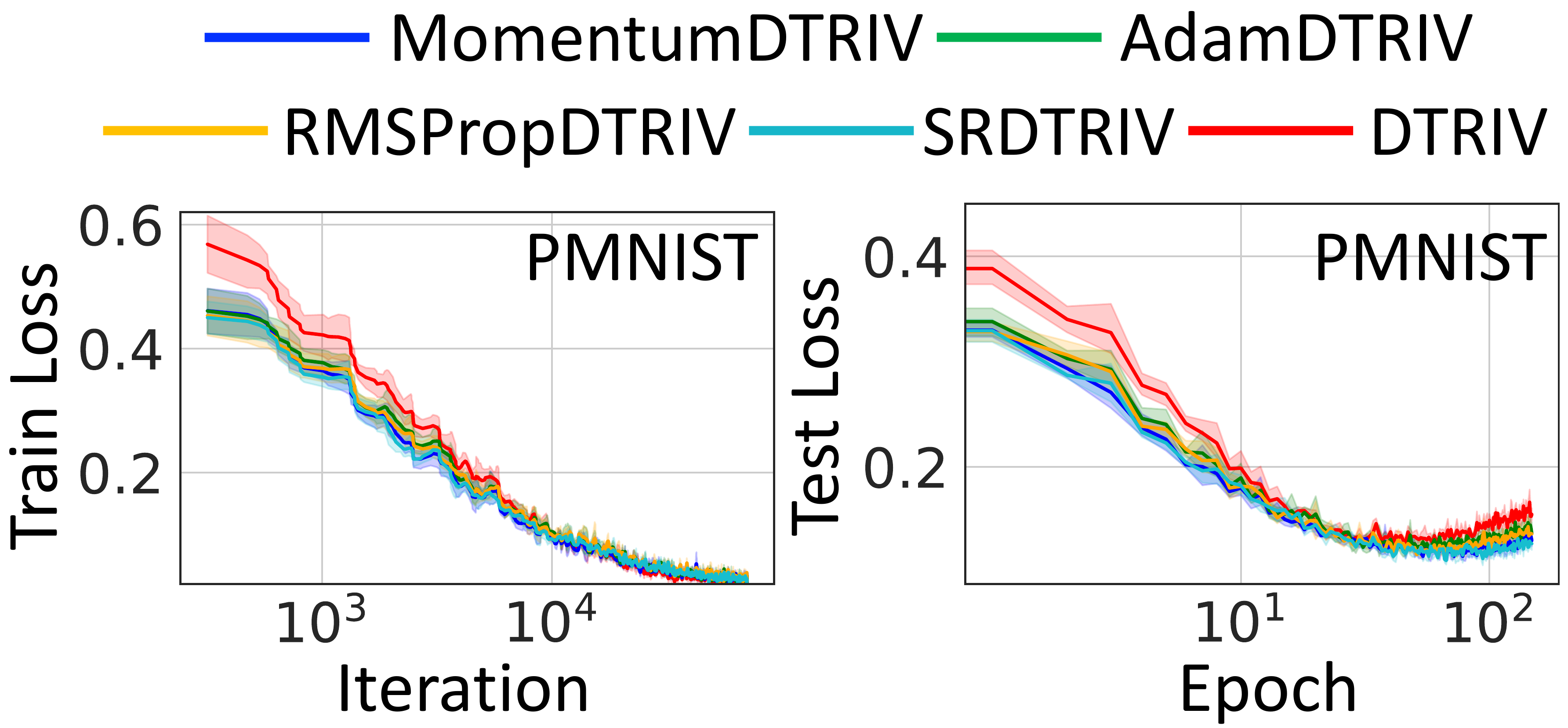}
\vskip -0.1cm
\caption{Train and test loss of MomentumDTRIV (blue), AdamDTRIV (green), RMSPropDTRIV (orange), SRDTRIV (cyan), and DTRIV (red) for PMNIST task. Our momentum-based models converge faster than the baseline DTRIV.}
\label{fig:loss-vs-iters-dtriv-appendix}
\end{figure}

Finally, we apply our Adam and NAG principled designing methods on a DTRIV, an orthogonal RNN~\cite{casado2019trivializations}, for the PMNIST classification task. We observe that AdamDTRIV, RMSPropDTRIV, and SRDTRIV outperform the baseline DTRIV while converging faster. SRDTRIV also outperforms MomentumDTRIV. We summarize our results in Table~\ref{tab:mnist-dtriv-appendix} and Figure~\ref{fig:loss-vs-iters-dtriv-appendix}. Hyperparameter values for this experiment can be found in Table~\ref{tab:alstm-hyperparams},~\ref{tab:rlstm-hyperparams}, and~\ref{tab:slstm-hyperparams} (bottom).

\begin{table}[!h]
    \caption{Best test accuracy at the MNIST and PMNIST tasks (\%). 
    We use the baseline results reported in \cite{pmlr-v80-helfrich18a},~\cite{wisdom2016full},~\cite{vorontsov2017orthogonality}. All of our proposed models outperform the baseline LSTM. Among the models using $N=128$ hidden units, RMSPropLSTM yields the best results in both tasks.}
\vspace{-0.1in}
\label{tab:mnist-appendix-n128}
\begin{center}
\begin{footnotesize}
\begin{sc}
\begin{tabular}{lclcc}
    \toprule
    Model & n & \# params & MNIST & PMNIST \\
    \midrule
    \midrule
    LSTM & $128$ & $\approx 68K$ & $98.70$\cite{pmlr-v80-helfrich18a},$97.30$  \cite{vorontsov2017orthogonality} & $92.00$  \cite{pmlr-v80-helfrich18a},$92.62$ \cite{vorontsov2017orthogonality} \\
    \midrule
    MomentumLSTM & $128$ & $\approx 68K$ & $\bf{99.04 \pm 0.04}$ & $\bf{93.40 \pm 0.25}$\\
    \midrule
    AdamLSTM & $128$ & $\approx 68K$ & $98.98 \pm 0.08$ & $93.75 \pm 0.25$\\
    RMSPropLSTM & $128$ & $\approx 68K$ & $\bf{99.09 \pm 0.05}$ & $\bf{94.32 \pm 0.43}$\\
    SRLSTM & $128$ & $\approx 68K$ & $98.89 \pm 0.08$ & $93.65 \pm 0.56$\\
    \bottomrule
\end{tabular}
\end{sc}
\end{footnotesize}
\end{center}
\end{table}

\begin{table}[!t]
    \caption{Test and validation MSEs at the end of the epoch with the lowest validation
MSE for the TIMIT task. All of our proposed models outperform the baseline LSTM. Among models using $N=120$ hidden units, MomentumLSTM performs the best.} 
\vspace{-0.2in}
\label{tab:timit-appendix-n120}
\begin{center}
\begin{footnotesize}
\begin{sc}
\begin{tabular}{lclcc}
    \toprule
    Model & n & \# params & Val. MSE & Test MSE \\
    \midrule
    \midrule
    LSTM & $120$ & $\approx 135K$ & $11.77 \pm 0.14$ ($13.93$ \cite{pmlr-v80-helfrich18a,lezcano2019cheap}) & $11.83 \pm 0.12$ ($12.95$ \cite{pmlr-v80-helfrich18a,lezcano2019cheap}) \\
    \midrule
    MomentumLSTM & $120$ & $\approx 135K$ & $\bf{8.00 \pm 0.30}$ & $\bf{8.04 \pm 0.30}$\\
    \midrule
    AdamLSTM & $120$ & $\approx 135K$ & $10.91 \pm 0.08$ & $10.96 \pm 0.08$\\
    RMSPropLSTM & $120$ & $\approx 135K$ & $11.83 \pm 0.20$ & $11.90 \pm 0.19$\\
    SRLSTM & $120$ & $\approx 135K$ & $8.15 \pm 0.26$ & $8.21 \pm 0.26$\\
    \bottomrule
\end{tabular}
\end{sc}
\end{footnotesize}
\end{center}
\end{table}

\begin{table}[!h]
\vspace{-0.15in}
    \caption{Best test accuracy on the PMNIST tasks (\%) for MomentumDTRIV and the baseline DTRIV, as well as for AdamDTRIV, RMSPropDTRIV, and SRDTRIV. We provide both our reproduced baseline results and those reported in~\cite{casado2019trivializations}. All of our momentum-based models outperform the baseline DTRIV. When using $N=512$ hidden units, SRDTRIV yields the best result.}
\vspace{-0.1in}
\label{tab:mnist-dtriv-appendix}
\begin{center}
\begin{small}
\begin{sc}
\begin{tabular}{lclc}
    \toprule
    Model & n & \# params & PMNIST \\
    \midrule
    \midrule
    DTRIV & $170$ & $\approx 16K$  & $95.21 \pm 0.10$ ($95.20$ \cite{casado2019trivializations}) \\
    DTRIV & $360$ & $\approx 69K$  & $96.45 \pm 0.10$ ($96.50$ \cite{casado2019trivializations}) \\
    DTRIV & $512$ & $\approx 137K$  & $96.62 \pm 0.12$ ($96.80$ \cite{casado2019trivializations}) \\
    \midrule
    MomentumDTRIV & $170$ & $\approx 16K$ & $\bf{95.37 \pm 0.09}$ \\
    MomentumDTRIV & $360$ & $\approx 69K$ & $\bf{96.73 \pm 0.08}$ \\
    MomentumDTRIV & $512$ & $\approx 137K$ & $\bf{96.89 \pm 0.08}$ \\
    \midrule
    AdamDTRIV & $512$ & $\approx 137K$ & $\bf{96.77 \pm 0.21}$ \\
    RMSPropDTRIV & $512$ & $\approx 137K$ & $\bf{96.75 \pm 0.12}$ \\
    SRDTRIV & $512$ & $\approx 137K$ & $\bf{97.02 \pm 0.09}$ \\
    \bottomrule
\end{tabular}
\end{sc}
\end{small}
\end{center}
\end{table}


\newpage
\section{Computational Time and Memory Cost: RNN vs. MomentumRNN}\label{sec:appendix:Time-and-Memory}

We provide the computation time and memory cost per sample at training and evaluation of MomentumLSTM, AdamLSTM, RMSPropLSTM, and SRLSTM in comparison with LSTM for PMNIST classification task using 256 hidden units in Table~\ref{tab:compute-time} and ~\ref{tab:memory-cost}, respectively.

\begin{table}[!h]
\vspace{-0.1in}
    \caption{Computation time per sample at training and evaluation for PMNIST classification task using models with 256 hidden units.}
\vspace{-0.1in}
\label{tab:compute-time}
\begin{center}
\begin{small}
\begin{sc}
\begin{tabular}{lcc}
    \toprule
    Model & Training Time ($\mu s$/sample) & Evaluation Time ($\mu s$/sample) \\
    \midrule
    \midrule
    LSTM & $6.18$ & $2.52$\\
    \midrule
    MomentumLSTM & $7.43$ & $3.16$\\
    AdamLSTM & $10.34$ & $4.07$\\
    RMSPropLSTM & $9.94$ & $3.96$\\
    SRLSTM & $8.34$ & $3.16$\\
    \bottomrule
\end{tabular}
\end{sc}
\end{small}
\end{center}
\end{table}

\begin{table}[!h]
\vspace{-0.1in}
    \caption{Memory cost per sample at training and evaluation for PMNIST classification task using models with 256 hidden units.}
\vspace{-0.1in}
\label{tab:memory-cost}
\begin{center}
\begin{small}
\begin{sc}
\begin{tabular}{lcc}
    \toprule
    Model & Training Memory (MB/sample) & Evaluation Memory (MB/sample) \\
    \midrule
    \midrule
    LSTM & $15.93$ & $7.51$\\
    \midrule
    MomentumLSTM & $15.95$ & $7.51$\\
    AdamLSTM & $25.13$ & $7.52$\\
    RMSPropLSTM & $25.13$ & $7.52$\\
    SRLSTM & $15.95$ & $7.51$\\
    \bottomrule
\end{tabular}
\end{sc}
\end{small}
\end{center}
\end{table}

\begin{table}[!t]
\vspace{-0.1in}
    \caption{Total computation time to reach the same 92.29\% test accuracy of LSTM (see Tab. 1) for PMNIST classification task using models with 256 hidden units.}
\vspace{-0.1in}
\label{tab:compute-time-to-reach-same-acc}
\begin{center}
\begin{small}
\begin{sc}
\begin{tabular}{lc}
    \toprule
    Model & Time ($min$) \\
    \midrule
    \midrule
    LSTM & $767$\\
    \midrule
    MomentumLSTM & $551$\\
    AdamLSTM & $\bf{225}$\\
    RMSPropLSTM & $416$\\
    SRLSTM & $348$\\
    \bottomrule
\end{tabular}
\end{sc}
\end{small}
\end{center}
\end{table}

\section{Additional Information about the Figures in the Main Text}

In Figure~\ref{fig:loss-vs-iters}, the MNIST plots are for models with 256 hidden units, and the TIMIT plots are for models with 158 hidden units.

In Figure~\ref{fig:loss-vs-iters-dtriv}, the PMNIST plots are for models with 512 hidden units, and the TIMIT plots are for models with 322 hidden units.
\vspace{0.2in}
\section{MomentumLSTM Cell Implementation in Pytorch}
\label{appendix:PyTorch-implementation}
\begin{lstlisting}[language=python]
import torch
import torch.nn as nn
from torch.nn import functional as F

class MomentumLSTMCell(nn.Module):

    """
    An implementation of MomentumLSTM Cell
    
    Args:
        input_size: The number of expected features in the input `x'
        hidden_size: The number of features in the hidden state `h'
        mu: momentum coefficient in MomentumLSTM Cell
        s: step size in MomentumLSTM Cell
        bias: If ``False'', then the layer does not use bias weights `b_ih' and `b_hh'. Default: ``True''

    Inputs: input, hidden0=(h_0, c_0), v0
        - input of shape `(batch, input_size)': tensor containing input features
        - h_0 of shape `(batch, hidden_size)': tensor containing the initial hidden state for each element in the batch.
        - c_0 of shape `(batch, hidden_size)': tensor containing the initial cell state for each element in the batch.
        - v0 of shape `(batch, hidden_size)': tensor containing the initial momentum state for each element in the batch

    Outputs: h1, (h_1, c_1), v1
        - h_1 of shape `(batch, hidden_size)': tensor containing the next hidden state for each element in the batch
        - c_1 of shape `(batch, hidden_size)': tensor containing the next cell state for each element in the batch
        - v_1 of shape `(batch, hidden_size)': tensor containing the next momentum state for each element in the batch
    """

    def __init__(self, input_size, hidden_size, mu, s, bias=True):
        super(MomentumLSTMCell, self).__init__()
        self.input_size = input_size
        self.hidden_size = hidden_size
        self.bias = bias
        self.x2h = nn.Linear(input_size, 4 * hidden_size, bias=bias)
        self.h2h = nn.Linear(hidden_size, 4 * hidden_size, bias=bias)
        
        # for momentumnet
        self.mu = mu
        self.s = s
        
        self.reset_parameters(hidden_size)

    def reset_parameters(self, hidden_size):
        nn.init.orthogonal_(self.x2h.weight)
        nn.init.eye_(self.h2h.weight)
        nn.init.zeros_(self.x2h.bias)
        self.x2h.bias.data[hidden_size:(2 * hidden_size)].fill_(1.0) 
        nn.init.zeros_(self.h2h.bias)
        self.h2h.bias.data[hidden_size:(2 * hidden_size)].fill_(1.0)
    
    def forward(self, x, hidden, v):
        
        hx, cx = hidden
        
        x = x.view(-1, x.size(1))
        v = v.view(-1, v.size(1))
        
        vy = self.mu * v + self.s * self.x2h(x)
        
        gates = vy + self.h2h(hx)
    
        gates = gates.squeeze()
        
        ingate, forgetgate, cellgate, outgate = gates.chunk(4, 1)
        
        ingate = F.sigmoid(ingate)
        forgetgate = F.sigmoid(forgetgate)
        cellgate = F.tanh(cellgate)
        outgate = F.sigmoid(outgate)
        
        cy = torch.mul(cx, forgetgate) +  torch.mul(ingate, cellgate)        

        hy = torch.mul(outgate, F.tanh(cy))
        
        return hy, (hy, cy), vy

\end{lstlisting}

\end{document}